\documentclass[10pt, journal, web]{article}
\usepackage{mathtools}
\usepackage{amssymb}
\usepackage{bbding}

\usepackage{listings}%
\usepackage{amssymb}

\usepackage{algorithmic}
\usepackage[linesnumbered,ruled,vlined]{algorithm2e}
\usepackage{epsfig}
\usepackage{multirow}
\usepackage{pifont}
\usepackage{graphicx}
\usepackage{xcolor}

\usepackage{multicol}
\usepackage{blindtext}
\usepackage{indentfirst}
\usepackage[hyphens]{url} 

\usepackage[colorlinks,
            linkcolor=red,
            anchorcolor=green,
            citecolor=green
            ]{hyperref}
\usepackage{lineno}

\usepackage[font=small,labelfont=bf]{caption}
\usepackage{subcaption}
\usepackage{chngpage}
\usepackage{caption}

\usepackage{fancyhdr,graphicx,amsmath,amssymb}

\usepackage{subcaption}

\usepackage{lipsum} %

\usepackage{tikz}
\usepackage{multirow}%
 \usepackage{amsmath,amssymb,amsfonts}%
 \usepackage{amsthm}%
\usepackage{mathrsfs}%
\usepackage[title]{appendix}%
\usepackage{xcolor}%
\usepackage{textcomp}%
\usepackage{manyfoot}%
\usepackage{booktabs}%
 \usepackage{listings}%

\usepackage{graphicx}
\usepackage{authblk}

\usepackage{longtable}
\usepackage{pdflscape}
\usepackage[T1]{fontenc}

\usepackage{xcolor}
\usepackage{hyperref}
\usepackage{booktabs}
\hypersetup{
   colorlinks=true,
    linkcolor=blue,
   citecolor=blue,
    urlcolor=blue
}
\usepackage{makecell}
\usepackage{tabularx}
\usepackage{enumitem}

\newcommand{\Xvect}{\mbox{\bf X}}

\DeclareUnicodeCharacter{2026}{\textellipsis}

\makeatletter
\renewcommand\AB@affilsepx{, \protect\Affilfont}
\makeatother
\usepackage{tabularx}
\providecommand{\keywords}[1]
{
  \small	
  \textbf{\textit{Keywords---}} #1
}

\begin{document}

\title{\textbf{HSMix: Hard and Soft Mixing Data Augmentation for Medical Image Segmentation}}
\author[1]{Danyang Sun}
\author[1, 2]{Fadi Dornaika\thanks{Corresponding author}}
\author[1]{Nagore Barrena}
\affil[1]{\textit{University of the Basque Country}}
\affil[2]{\textit{IKERBASQUE}}

\affil[ ]{

\small\texttt{danyangsun@163.com, fadi.dornaika@ehu.eus, nagore.barrena@ehu.eus}}
\date{}
\maketitle

\begin{abstract}
Due to the high cost of annotation or the rarity of some diseases, medical image segmentation is often limited by data scarcity and the resulting overfitting problem. Self-supervised learning and semi-supervised learning can mitigate the data scarcity challenge to some extent. However, both of these paradigms are complex and require either hand-crafted pretexts or well-defined pseudo-labels. In contrast, data augmentation represents a relatively simple and straightforward approach to addressing data scarcity issues. It has led to significant improvements in image recognition tasks. However, the effectiveness of local image editing augmentation techniques in the context of segmentation has been less explored. Additionally, traditional data augmentation methods for local image editing augmentation methods generally utilize square regions, \textcolor{black}{which cause} a loss of contour information.  
We propose HSMix, a novel approach to local image editing data augmentation involving hard and soft mixing for medical semantic segmentation. In our approach, a hard-augmented image is created by combining homogeneous regions (superpixels) from two source images. A soft mixing method further adjusts the brightness of these composed regions with brightness mixing based on locally aggregated pixel-wise saliency coefficients. The ground-truth segmentation masks of the two source images undergo the same mixing operations to generate the associated masks for the augmented images. Our method fully exploits both the prior contour and saliency information, thus preserving local semantic information in the augmented images while enriching the augmentation space with more diversity. Our method is a plug-and-play solution that is model agnostic and applicable to a range of medical imaging modalities. Extensive experimental evidence has demonstrated its effectiveness in a variety of medical segmentation tasks. The source code is available in \url{https://github.com/DanielaPlusPlus/HSMix}.

\end{abstract}

 \hspace{10pt}

\section{Introduction}

Medical and biological image analysis has been significantly advanced by deep learning \cite{xu2021emfusion,jha2020doubleu}. However, deep learning models are inherently prone to overfitting with limited training images in medical visual tasks.
 Medical image segmentation is a crucial process that involves identifying and delineating specific regions for further diagnosis in the context of computer-aided diagnosis (CAD) systems \cite{qiu2022dwarfism}. The collection and annotation of medical training images for medical image segmentation can be costly, impractical, or require the involvement of special cases and professional expertise.  In summary, data scarcity and overfitting issues can seriously limit the medical image segmentation process \cite{ma2024segment}. Most work on medical image segmentation focused on novel deep architectures \cite{khan2021deep, jha2020doubleu, jha2019resunet++}. 
 Nevertheless, less attention has been paid to dealing with data scarcity through model-agnostic strategies.
 Researchers often use either self-supervised learning \cite{zhang2023dive} or semi-supervised learning \cite{jiao2023learning} to address the lack of labeled data.
 Both approaches require the use of external data, even if these are not labeled. However, access to medical data is challenging due to privacy and legal concerns. In addition, self-supervised learning requires carefully designed replacement tasks, which are very difficult to define in the context of medical image segmentation.

Data augmentation \cite{wang2021regularizing} can address data scarcity and overfitting problems by generating more training samples \cite{goceri2023medical}. In contrast to semi-supervised learning and self-supervised learning, data augmentation can be used to improve performance more directly. Basic transformation techniques and generative adversarial networks (GANs) \cite{goodfellow2020generative} are commonly used data augmentation methods for  medical image segmentation tasks. However, basic transformation approaches such as flipping, cropping, and rotation have limited effectiveness. Besides, the application of  GANs to data augmentation is computationally expensive and presents significant challenges in training using an adversarial framework. 
 
It has been shown that the implementation of local image editing augmentation methods improves the generalization ability of the deep model. This is achieved by randomly directing the model's attention to different local regions by randomly removing or occluding data during the augmentation process. However, most of these methods have been proposed and evaluated for image recognition, such as Mixup \cite{zhang2018mixup}, CutOut \cite{devries2017improved}, CutMix \cite{yun2019cutmix} and  SuperpixelGridMix \cite{hammoudi2022superpixelgridmasks}.
On the other hand, for segmentation tasks, only limited research has been conducted using local image editing augmentation. Moreover, conventional local image editing data augmentation methods often use square regions, which is suboptimal given the loss of contours.

Inspired by the classic Mixup \cite{zhang2018mixup} and CutMix \cite{yun2019cutmix}, a novel approach to local image editing augmentation is proposed using superpixels, which includes a hard and a soft mixing.The hard mixing component uses a superpixel-based CutMix and preserves the contour information in the augmented images, while the soft mixing component uses a superpixel-based Mixup with saliency and provides soft augmented images and segmentation masks  for robust boosts. \textcolor{black}{The} hard augmented image is created by combining homogeneous regions (superpixels) from two source images. \textcolor{black}{The} soft-mixing method adjusts the brightness of these composite regions with brightness mixing based on locally aggregated pixel-wise saliency coefficients. The segmentation mask of the two source images is subjected to the same blending operations to generate the corresponding mask for the augmented image. The hard and soft augmentation together expand a more diverse space for both the images and segmentation masks.

\begin{figure*}[t]
	\centering
	\includegraphics[scale=0.65]{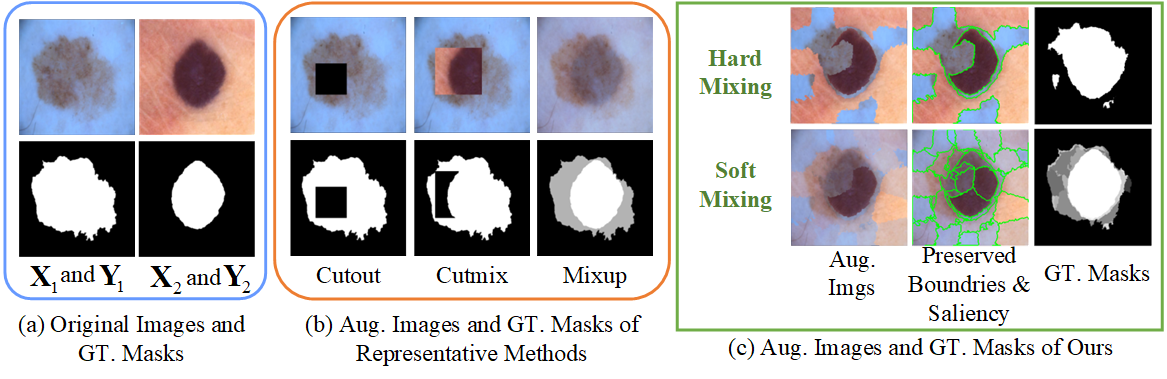}
    \caption{Visualization of the augmentation (augmented images and corresponding GT. masks) from representative methods. Our method in (c) can preserve more boundaries and saliency information in a more diverse space. \textcolor{black}{$\mathbf{X}_{1}$ and $\mathbf{X}_{2}$ denote two original images, and $\mathbf{Y}_{1}$ and $\mathbf{Y}_{2}$ denote the corresponding ground truth masks.}}
    \label{fig1}
\end{figure*}

Figure \ref{fig1} presents a comparison of augmentation derived from disparate local image editing augmentation approaches. 
Figure \ref{fig1}(a) illustrates two arbitrary training samples and \textcolor{black}{their} respective ground truth masks. Figure \ref{fig1}(b) represents how CutOut \cite{devries2017improved} is applied: an aleatory squared is removed (setting its pixel values to zero) from the image. Contour information is missing using that method. 
The CutMix method \cite{yun2019cutmix} involves randomly cutting a \textcolor{black}{square} region from one image and pasting it onto another image. Similarly, this process also loses contours. Mixup \cite{zhang2018mixup} blends the images and labels pixel-to-pixel with a randomly generated proportion that is fixed throughout the process. Our method in Figure \ref{fig1}(c) performs both hard mixing (regional mixing with cut-and-paste) and soft mixing (pixel-to-pixel mixing) with superpixel grids, respectively. The former allows for the preservation of contour information, while the latter utilizes the related saliency maps of the two images as a blending proportion in every superpixel, rather than a fixed value for all pixels. This ensures the preservation of both contour and salient information in the augmented images and masks. The combination of hard and soft mixing results in a more diverse set of augmented images and their corresponding labels.

In comparison to either self-supervised learning or semi-supervised learning, our method is relatively straightforward with only one stage. In contrast to basic transformation data augmentation methods, our approach is more effective. In contrast to GAN-model-based augmentation approaches, the method presented is independent of the model and easy to implement. In contrast to existing other local image editing augmentation methods, ours maintains contour information in the augmented images and creates an augmentation space with greater diversity. 

Our method contributes in threefold:
\begin{enumerate}
\item We present an easy and effective technique for augmenting medical segmentation data. To the best of  our knowledge, this is the first hard and soft mixing-based data augmentation technique that fully leverages both prior contour and saliency information.

\item The augmentation space is broadened with greater diversity through the integration of hard and soft mixing, which is crucial for image and segmentation mask augmentation.

\item Our method is a plug-and-play solution that works with any model and can be used with different medical imaging modalities. In numerous experiments conducted with different medical datasets and image types, it has been shown to be superior to other methods.

\end{enumerate}





\section{Related Work}
\label{related}

In this section, we provide an overview of the most important work in the areas relevant to our research. First, we will examine studies dealing with medical image segmentation. Next, we will review research that deals with the use of data augmentation techniques specifically for segmentation tasks in medical images. \textcolor{black}{Finally}, we will discuss recent work investigating the use of superpixel techniques for data augmentation.


\subsection{Segmentation task in medical images}

The segmentation task is an active area of research due to its importance for a variety of clinical and research applications. In recent years, numerous methods and techniques have been developed to address this challenge. Advances in deep learning algorithms, image processing techniques\textcolor{black}{,} and the availability of annotated biomedical datasets have been utilized. Common tasks in medical image segmentation include the segmentation of the liver and liver tumors \cite{li2015automatic, vivanti2015automatic}, the brain and brain tumors \cite{menze2014multimodal, cherukuri2017learning}, optic nerve head \cite{cheng2013superpixel, fu2018joint}, cell \cite{song2017dual, wei2020mitoem}, lung nodules \cite{wang2017central}, heart images \cite{wu2020cf}, \textcolor{black}{and} images of skin lesions \cite{yuan2017automatic}, among others. In particular, deep learning models have shown considerable potential in the segmentation of medical images \cite{litjens2017survey, minaee2021image}. They are able to capture complicated image features and provide precise segmentation results for various tasks, from the delineation of specific anatomical structures to the detection of pathological areas \cite{antonelli2022medical}.

In this way, the use of Convolutional Neural Networks (CNNs) in the segmentation of medical images has evolved rapidly. CNNs have shown remarkable success in these tasks due to their ability to learn hierarchical features from data. \textcolor{black}{The} U-Net architecture proposed by Ronneberger et al. \cite{ronneberger2015u} is still one of the most popular solutions for medical image segmentation tasks. Its symmetric encoder-decoder architecture with skip connections helps to preserve the spatial information, while the network can capture both local and global contexts. Numerous variations and improvements to the original U-Net architecture have been proposed to address specific challenges in medical image segmentation tasks. These variations include Attention U-Net \cite{oktay2018attention}, V-Net \cite{milletari2016v}, and Residual U-Net \cite{zhang2018road}, among others. They include attention mechanisms, residual connections, or other modifications to improve segmentation performance.

Transformers have shown promise in various natural language processing tasks, and their application to computer vision tasks, including medical image segmentation, is gaining traction \cite{lin2022survey}. While CNNs have been the predominant choice for image segmentation tasks, Transformers offer a different approach to capturing spatial relationships and contextual information in images. Recent advances in Transformer architectures, such as the Vision Transformer (ViT) and its variants, have shown to perform competitively in image classification tasks by efficiently processing image patches without the need for explicit convolutional operations \cite{bazi2021vision}. These architectures can be adapted for image segmentation by integrating additional layers or modules for segmentation-specific tasks.

However, the use of CNNs and transformers for medical image segmentation depends on the scarcity and quality of the annotated data\textcolor{black}{,} as well as the interpretability and generalization of the models. Annotating medical images for segmentation tasks is often a time-consuming and labor-intensive process that requires expert knowledge and manual delineation of regions of interest. This leads to a limited availability of annotated datasets. CNNs trained on limited annotated data are prone to overfitting, where the model memorizes noise or peculiarities in the training data instead of learning meaningful representations. Similarly, when there is a lack of annotated medical images, it becomes difficult to train transformer-based segmentation models that generalize well to unseen data.

Either self-supervised learning or semi-supervised learning is often utilized to mitigate the shortage of labeled data.  

For example, Xie et al. \cite{xie2023maester} present a self-supervised method for accurate, sub-cellular structure segmentation at pixel resolution. Numerous works are presented in the state of the art which tackle medical image segmentation tasks, using either self-supervised learning (\cite{tang2022self, xie2020pgl, chaitanya2023local, shurrab2022self,ouyang2022self}) or semi-supervised learning (\cite{luo2021semi, bortsova2019semi, chen2019multi, bai2017semi, sun2024semi}).

However, obtaining additional unannotated data for certain medical tasks is difficult due to its rarity and privacy concerns. In addition, methods based on self-supervised learning involve two steps in the training process: pre-training and fine-tuning. It is important to ensure the potential of transfer learning. This is possible by designing a very effective and well-performing pre-text task in the pre-training phase. However, it should be noted that in many cases good performance in the pre-text task leads to poor performance in subsequent tasks \cite{xie2022simmim}. In addition, the use of image transformations or processing to create self-labeled data or pre-texts is essential to improve feature representation capabilities. This is also crucial in semi-supervised learning \cite{bai2023bidirectional, haghighi2024self, weng2024semi}.

\subsection{Data Augmentation in medical image segmentation}

Addressing the lack of labeled data in medical image segmentation requires creative approaches and efficient use of available resources. Data augmentation methods have been developed to solve this problem. They artificially expand the labeled dataset. Data augmentation can bring diversity to the training data and improve the model's ability to generalization \cite{goceri2023medical}.

Many previous works in the field of medical image segmentation have used data augmentation techniques applying basic transformations such as the following: Image rotation \cite{chaitanya2021semi}, image inversion \cite{hussain2017differential}, shifting the image \cite{sharma2017automatic, alnazer2021recent} and even cropping the image \cite{khan2021brain}.

The use of GANs is also widespread in the field of medical imaging \cite{abdelhalim2021data, ding2021high}. GAN-based techniques for data augmentation involve training both a generator and a discriminator. The state of the art includes some medical image analysis where these data augmentation techniques are used, \textcolor{black}{such as liver lesion} \cite{frid2018gan}, brain (intracranial hemorrhage) \cite{toikkanen2021resgan}, brain tumor \cite{han2019combining}, skin lesion \cite{lei2020skin}\textcolor{black}{, and} neuronal images \cite{zhao2018synthesizing} among others. However, GAN-based methods for data augmentation are often difficult to train and very computationally intensive.

Many local image editing augmentation methods
have proved their generalization ability in image classification, such as Mixup \cite{zhang2018mixup}, CutOut \cite{devries2017improved}, and CutMix \cite{yun2019cutmix}. They are also applied in medical image segmentation, such as brain tumor segmentation \cite{wang2022data}, abdomen segmentation \cite{zhang2022self}\textcolor{black}{, and} liver tumor \cite{basaran2023lesionmix} among others. 

The effectiveness of these augmentation techniques has not been thoroughly investigated in the context of segmentation tasks. We propose a method for local image processing that includes both hard and soft blending with superpixel grids.

\subsection{Superpixels for Data Augmentation}
Superpixels represent an over-segmentation technique that combines neighboring pixels with similar low-level visual properties. The most commonly used algorithm for creating superpixels is the so-called Simple Linear Iterative Clustering (SLIC) \cite{achanta2012slic}.

The manipulation of superpixels can lead to a remarkable reduction in computational effort compared to pixel-wise methods, while ensuring the preservation of image boundaries and contours. The granularity of superpixels can be varied to capture either the broad or the intricate features of the original images.


Due to their ability to efficiently represent images and perform computations effectively, superpixels have been studied and used in deep learning models for visual tasks.
Our main interest is the use of superpixels for data augmentation. While they have been studied for classification tasks, their use for data augmentation for semantic image segmentation is still relatively unexplored. Wang et al. \cite{wang2023autosmim} \textcolor{black}{presented} autoSMIM, a method for segmenting medical images (especially skin lesions) using superpixels. However, they \textcolor{black}{did} not use superpixels for data augmentation. It is a self-supervised method that uses the reconstruction of the removed superpixel as a default. Looking at the methods of superpixel-based data expansion, the following outlines are used, but for classification. The approach proposed in \cite{accion2020dual} involves the extraction of patches from the superpixels, which are then subjected to geometric transformations. Hammoudi et al \cite{hammoudi2022superpixelgridmasks} \textcolor{black}{proposed} randomly hiding, erasing, or mixing along the superpixel grids for image classification tasks. OcCaMix 
\cite{dornaika2023object} extends the work of  \cite{hammoudi2022superpixelgridmasks} and   performs superpixel mixing guided by attention. LGCOAMix \cite{dornaika2023lgcoamix} performs random superpixel mixing  while generating the corresponding mixed labels with attention. SAFusion \cite{sun2024semi} proposes an augmentation strategy for images and corresponding labels that is based on superpixel-based fusion. The aforementioned works are designed for classification tasks. 

The following two works are intended for segmentation tasks. Franchi et al. \cite{franchi2021robust} \textcolor{black}{performed} random superpixel mixing for semi-supervised semantic segmentation. The method in \cite{zhang2019spda} \textcolor{black} {proposed} to use the superpixelized image directly as the augmented image for segmentation. 

To summarize, the potential of data augmentation with superpixels for semantic segmentation is still largely untapped. Our approach presents a novel augmentation method for medical semantic segmentation tasks using hard and soft pairwise image blending with superpixel grids.


\textcolor{black}{A review of the existing literature on medical image segmentation reveals the following limitations.}



\begin{itemize}
\item \textcolor{black}{Segmentation tasks in medical imaging often result in overfitting, largely due to the scarcity of annotated data.}

\item \textcolor{black}{Self-supervised and semi-supervised approaches can be of some help, but are inherently complex. In addition, obtaining unlabeled data in the medical field remains a costly endeavor.}

\item \textcolor{black}{Most conventional data augmentation techniques are not contour sensitive, which leads to the loss of important boundary information.}

\item \textcolor{black}{The use of GAN models for data augmentation has proven to be an effective approach, but is a complex and time-consuming method.}

\item Superpixels bundle pixels with similar visual properties and preserve contour information. Their use in data augmentation is being explored for classification tasks, but not for segmentation.

\end {itemize}


So, in order to avoid \textcolor{black}{the} mentioned limitations, this work introduces an innovative  model-agnostic technique for local image editing and data augmentation tailored for the semantic segmentation of medical images. This approach leverages both hard and soft mixing, making full use of superpixel regions and pre-existing saliency information. By doing so, it maintains critical contour details in the augmented images while enhancing the diversity of the augmentation space.

\section{Methodology}
\label{methodology}


In our method, two training images in a given batch, $\Xvect_1$ and  $\Xvect_2$\textcolor{black}{,} are simultaneously subjected to the hard mixing augmentation (Sec. \ref{hard-mixer}) and the soft mixing augmentation (Sec. \ref{soft-mixer}). In other words, each fusion creates a hard and a soft augmentation image. Subsequently, both the hard and soft augmented images and \textcolor{black}{their} corresponding ground truth masks are used \textcolor{black}{to train} a deep segmentation model (Sec. \ref{train-infer}).

\subsection{Hard Mixing with Superpixel-based Cutmix}
\label{hard-mixer}

\begin{figure*}[h]
	\centering
	\includegraphics[scale=0.51]{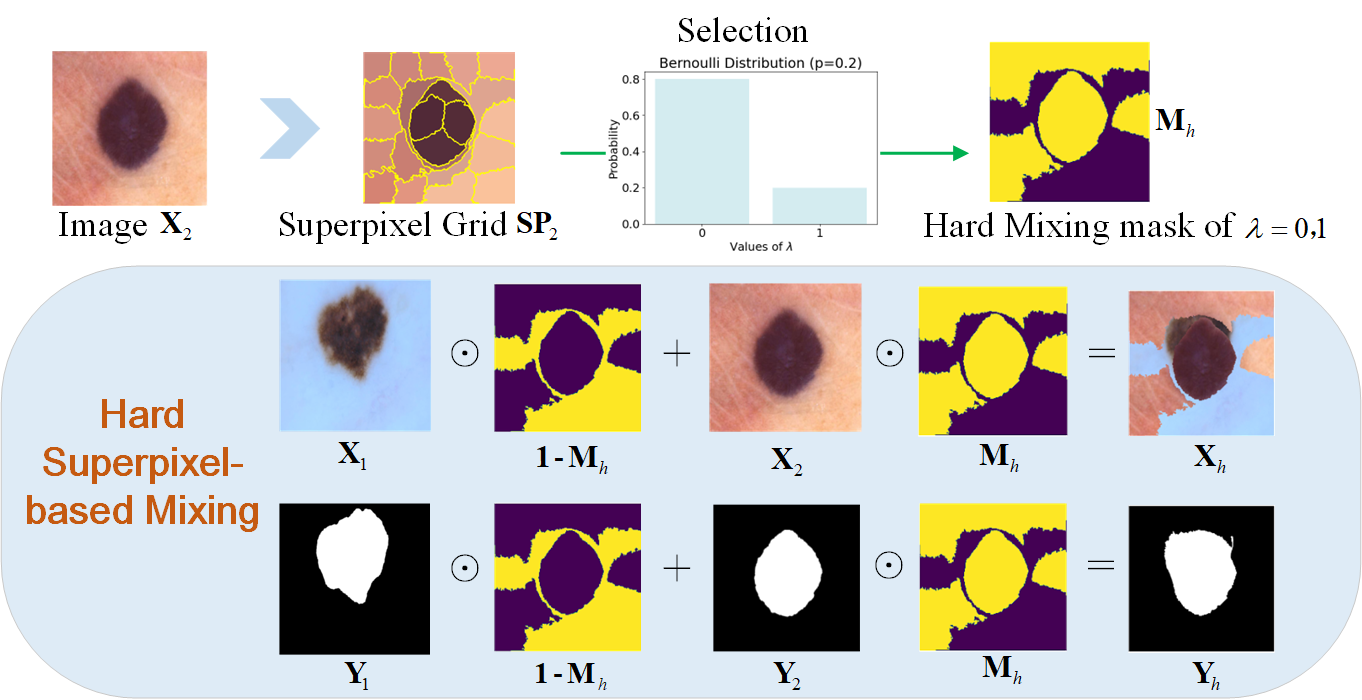}
    \caption{The process of hard mixing. \textcolor{black}{$\mathbf{X}_{1}$, $\mathbf{X}_{2}$ and $\mathbf{Y}_{1}$,  $\mathbf{Y}_{2}$ denote the two original images and their corresponding ground truth masks. $\mathbf{X}_{h}$ is the augmented image, $\mathbf{Y}_{h}$ is the ground truth mask of the augmented image. $\mathbf{M}_{h}$ is the binary mask for hard mixing.}}
    \label{fig2}
\end{figure*}

In hard mixing, the pairwise images are mixed with a mixing proportion, denoted by $\lambda$, which can be 0 or 1. This process is therefore referred to as hard mixing. \textcolor{black}{In the process of hard mixing, the samples are augmented by superpixel-based cutmix (cut-and-paste)}. In contrast to traditional cutmix methods \cite{yun2019cutmix}, our approach involves the use of superpixel regions instead of \textcolor{black}{square} regions. \textcolor{black}{This enables the retention of contour information within the augmented images. The augmented image with enhanced contour information can facilitate more effective contour recognition and more accurate segmentation through model training.} As indicated in Figure \ref{fig2} and Algorithm \ref{algo1} (see lines 4-6), the augmented image, denoted by \textcolor{black}{$\mathbf{X}_{h}$}, and its corresponding segmentation mask, \textcolor{black}{$\mathbf{Y}_{h}$}, are generated through a superpixel-based cutmix process. In particular, the superpixel grid map $\mathbf{SP}_{2}$ of the image $\mathbf{X}_{2}$ is employed to select each superpixel in a Bernoulli distribution, as described in Eq. (\ref{eq1}).

\begin{equation}\label{eq1}
P\{Z=k\}=p^k(1-p)^{1-k}, Z\sim{B(1,p)}, k=0,1
\end{equation}
where $Z$ denotes the random selection process of superpixels based on the superpixel grid $\mathbf{SP}_{2}$ of image $\mathbf{X}_{2}$. $B$ denotes the Bernoulli distribution with parameter $p$.


If a superpixel is selected in \textcolor{black}{the} image $\Xvect_2$ (i.e., it will be used in the augmented image), the mixing proportion $\lambda$ is set to 1, otherwise it is set to 0. This results in the generation of the hard mixing mask $\mathbf{M}_{h}$ with mixing proportion of either 0 or 1. With the hard mixing mask $\mathbf{M}_{h}$, we generate the augmented image \textcolor{black}{$\mathbf{X}_{h}$} and its corresponding label \textcolor{black}{$\mathbf{Y}_{h}$}, as described in Eq. (\ref{eq2}) and Figure \ref{fig2}.

\begin{equation}\label{eq2}
\begin{array}{c}
    \textcolor{black}{\mathbf{X}_{h}} = {\mathbf{X}_1}\odot(\mathbf{1}-\mathbf{M}_{h}) + {\mathbf{X}_2}\odot\mathbf{M}_{h}\\
    \textcolor{black}{\mathbf{Y}_{h}} = {\mathbf{Y}_1}\odot(\mathbf{1}-\mathbf{M}_{h}) + {\mathbf{Y}_2}\odot\mathbf{M}_{h}\\
\end{array}
\end{equation}
where $\mathbf{X}_1$, $\mathbf{X}_2$ and $\mathbf{Y}_1$, $\mathbf{Y}_2$ denote the original images and their ground truth segmentation masks. $\mathbf{1}$ represents the mask with the values one. $\mathbf{M}_{h}$ is a binary mask for mixing with \textcolor{black}{the values either 0 or 1}. All have the size $W \times H$. $\odot$ denotes the element-wise multiplication.

\subsection{Soft Mixing with Superpixel-saliency-based Mixup}
\label{soft-mixer}

In hard mixing, the resulting augmented image combines the local superpixels without mixing the channel dimension (color or brightness).
In soft mixing, the paired images are mixed with an adaptive mixing ratio, \textcolor{black}{$\lambda$}, between 0 and 1. This process is therefore referred to as soft mixing. As shown in Algorithm \ref{algo1} (see lines 7-13), the images are augmented by a superpixel-based mixup during soft mixing.

\begin{figure*}[h]
	\centering
	\includegraphics[scale=0.54]{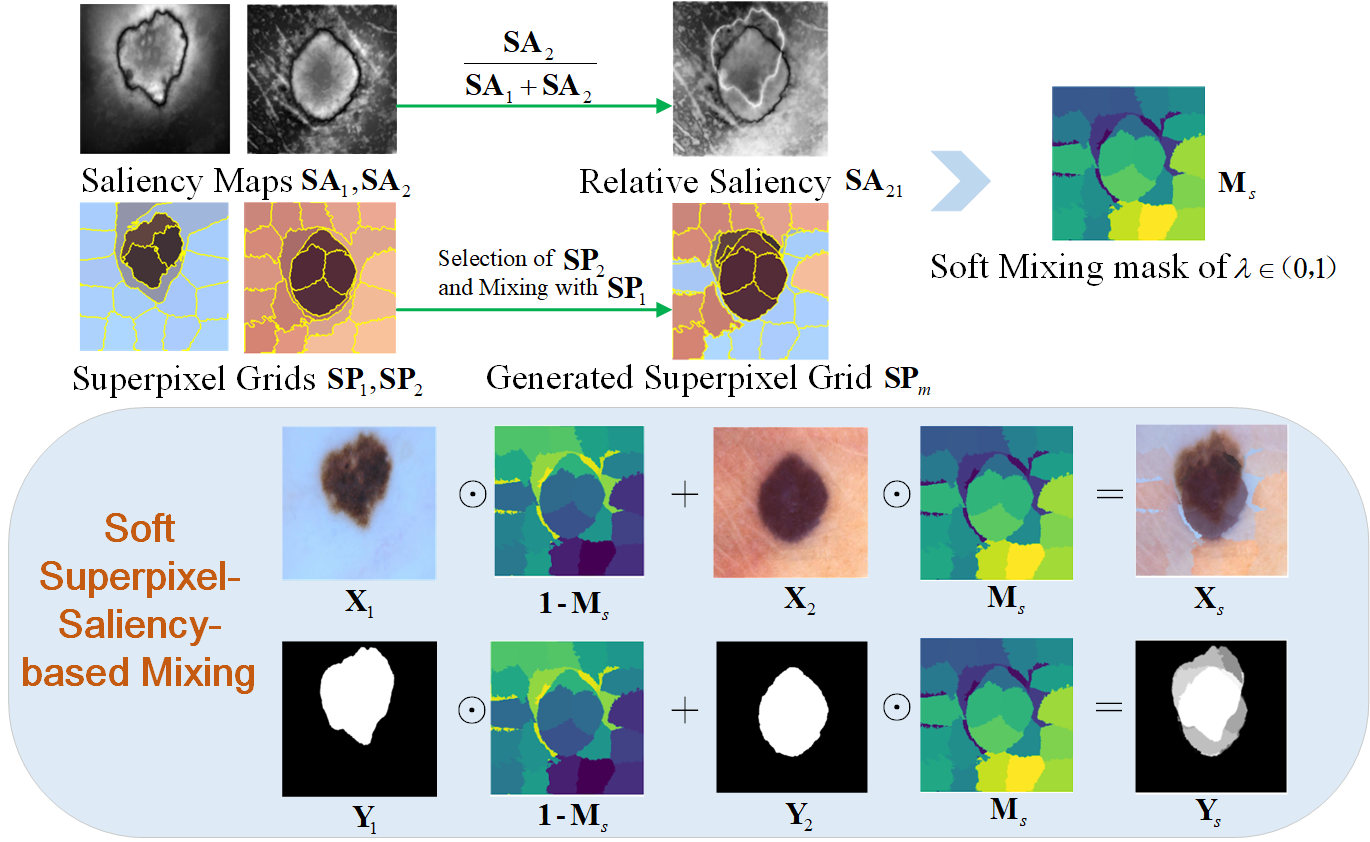}
    \caption{The process of soft mixing. \textcolor{black}{$\mathbf{X}_{1}$, $\mathbf{X}_{2}$ and $\mathbf{Y}_{1}$,  $\mathbf{Y}_{2}$ denote the two original images and their corresponding ground truth masks. $\mathbf{X}_{s}$ is the augmented image, $\mathbf{Y}_{s}$ is the ground truth mask of the augmented image. $\mathbf{M}_{s}$ is the mask for soft mixing.}}
    \label{fig3}
\end{figure*}


Given $\mathbf{SA}_{1}$ and $\mathbf{SA}_{2}$ are the normalized saliency maps of images $\mathbf{X}_{1}$ and $\mathbf{X}_{2}$. $\mathbf{SP}_{1}$ and $\mathbf{SP}_{2}$ are the superpixel grids of images $\mathbf{X}_{1}$ and $\mathbf{X}_{2}$. We generate the relative saliency map $\mathbf{SA}_{21}$ \textcolor{black}{of image $\mathbf{X}_{2}$} as described in Eq. (\ref{eq3}).



\begin{equation}\label{eq3}
    \mathbf{SA}_{{21}_{(i,j)}} =\frac{\mathbf{SA}_{2_{(i,j)}}}{\mathbf{SA}_{1_{(i,j)}}+\mathbf{SA}_{2_{(i,j)}}},
\end{equation}
where $(i,j)$ means the pixel of the $i_{th}$ row and the $j_{th}$ column in the $W\times{H}$ map. Using a similar formula,  we can get the relative saliency map of $\Xvect_1$ ($ \mathbf{SA}_{{12}_{(i,j)}}$) as  $\mathbf{SA}_{{12}_{(i,j)}} = 1-    \mathbf{SA}_{{21}_{(i,j)}}$. \textcolor{black}{The employment of saliency information signifies the structural information of the image, thereby directing attention for the deep model and capitalizing on the intrinsic structure and hierarchy of feature presentation. The utilization of relative saliency in lieu of the saliency of a single image can serve to enhance the robustness of our method.}
We emphasize that there are many efficient tools in the image processing literature to obtain the saliency map.

The mixed superpixel grid $\mathbf{SP}_{m}$ can be obtained with the hard mixing mask $\mathbf{M}_{h}$, as shown in Eq. (\ref{eq4}).

\begin{equation}\label{eq4}
    \mathbf{SP}_{m} = \textcolor{black}{{\mathbf{SP}_1}}\odot(\mathbf{1}-\mathbf{M}_{h}) + \textcolor{black}{{\mathbf{SP}_2}}\odot\mathbf{M}_{h}
\end{equation}
Here $\mathbf{SP}_{m}$ plays the role of \textcolor{black}{the} superpixel grid of the augmented image. Within a given superpixel in $ \mathbf{SP}_{m}$, the pixels are assigned the same mixing fraction by averaging the corresponding saliency over the entire superpixel as described in Eq.  (\ref{eq5}). The soft mixing mask, denoted by $\mathbf{M}_{s}$, with mixing fractions between 0 and 1, is described in Eq. (\ref{eq6}).

\begin{equation}\label{eq5}
    \lambda_{k} = \frac{1}{N_{k}}{\sum_{(i,j)\in{\Gamma_{k}}}^{N_{k}}}\mathbf{SA}_{{21}_{(i,j)}}, \:  \: \: \: k=1,2,3...,L 
\end{equation}

\begin{equation}\label{eq6}
    \mathbf{M}_{{S}_{(i,j)}} = \lambda_{k}, \: \: \: \:  (i,j)\in{\Gamma_{k}}
\end{equation}
where $\Gamma_{k}$ denotes the $k_{th}$ superpixel in the mixed superpixel grid $\mathbf{SP}_{m}$ generated in Eq. (\ref{eq4}). $\mathbf{SA}_{21}$ presents the relative  saliency map of the image $\Xvect_2$ (Eq. (\ref{eq3})). \textcolor{black}{For}  $L$ superpixels in $\mathbf{SP}_{m}$, the number of pixels in the $k_{th}$ superpixel is $N_{k}$. $\lambda_{k}$ is the mixing proportion for the $k_{th}$ superpixel $\Gamma_{k}$. 

With the soft mixing mask $\mathbf{M}_{s}$, we generate the augmented image \textcolor{black}{$\mathbf{X}_{s}$} and its corresponding label \textcolor{black}{$\mathbf{Y}_{s}$}, as described in Eq. (\ref{eq7}) and Figure \ref{fig3}.

\begin{equation}\label{eq7}
\begin{array}{c}
    \textcolor{black}{\mathbf{X}_{s}} = {\mathbf{X}_1}\odot(\mathbf{1}-\mathbf{M}_{s}) + {\mathbf{X}_2}\odot\mathbf{M}_{s}\\
    \textcolor{black}{\mathbf{Y}_{s}} = {\mathbf{Y}_1}\odot(\mathbf{1}-\mathbf{M}_{s}) + {\mathbf{Y}_2}\odot\mathbf{M}_{s}\\
\end{array}
\end{equation}

\begin{algorithm}[htbp] 
\SetKwFunction{Union}{Union}\SetKwFunction{FindCompress}{FindCompress} \SetKwInOut{Input}{Input}\SetKwInOut{Output}{Output}

	\Input{Two arbitrary images $\mathbf{X}_{1}$, $\mathbf{X}_{2}$ and their ground truth masks $\mathbf{Y}_{1}$, $\mathbf{Y}_{2}$; Number of superpixels range $l_{min}$, $l_{max}$; Superpixel selection probability $p$}
	
	\Output{Augmented images \textcolor{black}{($\mathbf{X}_{h}$,$\mathbf{X}_{s}$)} in a batch  and their corresponding ground truth masks \textcolor{black}{($\mathbf{Y}_{h}$,$\mathbf{Y}_{s}$)}}
    \BlankLine 
    \BlankLine 
  {
  {$l_{1}\sim {U}(l_{min},l_{max})$, $l_{2}\sim {U}(l_{min},l_{max})$}\\
  Superpixel grid $\mathbf{SP}_{1}$ $\leftarrow$ Superpixel algorithm($\Xvect_1$,$l_1$)\\
  Superpixel grid $\mathbf{SP}_{2}$ $\leftarrow$ Superpixel algorithm($\Xvect_2$,$l_2$)\\   
  \tcc{Hard mixing augmentation} 
  {Z  $\leftarrow$ Random selection with Eq. (\ref{eq1})}\\
    $\mathbf{M}_{h}  \in\mathbb{B}^{W\times{H}}$  $\leftarrow$ $ Ind (Select(\mathbf{SP}_2, Z))$ \\
  Generate \textcolor{black}{$\mathbf{X}_{h}$,$\mathbf{Y}_{h}$} from $\mathbf{X}_{1}$, $\mathbf{X}_{2}$,$\mathbf{Y}_{1}$, $\mathbf{Y}_{2}$ using Eq. (\ref{eq2}) and $\mathbf{M}_{h}$  \\ 
    \tcc{Soft mixing augmentation} 
  Saliency map $\mathbf{SA}_{1}$ $\leftarrow$ Saliency algorithm($\Xvect_1$)\\
  Saliency map $\mathbf{SA}_{2}$ $\leftarrow$ Saliency algorithm($\Xvect_2$)\\
  Calculate $\mathbf{SA}_{21}$ from   $\mathbf{SA}_{1}$,  $\mathbf{SA}_{2}$ using  Eq. (\ref{eq3})\\
  Generate $\mathbf{SP}_{m}$ from \textcolor{black}{$\mathbf{SP}_{1}$}, \textcolor{black}{$\mathbf{SP}_{2}$} using Eq. (\ref{eq4}) and $\mathbf{M}_{h}$\\ 
  Calculate $\lambda$ for every superpixel in $\mathbf{SP}_{m}$  using $\mathbf{SA}_{21}$ and  Eq. (\ref{eq5})\\ 
  Determine $\mathbf{M}_{s}$ using $\lambda$ of every superpixel along $\mathbf{SP}_{m}$ with Eq. (\ref{eq6})\\ 
  Generate \textcolor{black}{$\mathbf{X}_{s}$,$\mathbf{Y}_{s}$} from $\mathbf{X}_{1}$, $\mathbf{X}_{2}$,$\mathbf{Y}_{1}$, $\mathbf{Y}_{2}$ using Eq. (\ref{eq7})
  }
\caption{\textbf{HSMix}
}
\label{algo1}
\end{algorithm}

\subsection{Training and Inference}
\label{train-infer}

During training, each random pair of images can generate two augmented images. These two images are generated \textcolor{black}{by} \textcolor{black}{the} hard mixing and \textcolor{black}{the} soft mixing. Both are used for training, thereby increasing the diversity of the training data sets. It is \textcolor{black}{noteworthy} that both the hard and soft mixing techniques are applied to the same superpixel grids. They \textcolor{black}{reinforce each other} with the hard mixing providing \textcolor{black}{accurate labeling} and the soft mixing providing \textcolor{black}{smooth labeling} in the augmented mask.  The training is conducted using the common dice loss and the cross-entropy loss, as described in Eq. (\ref{eq8}). The data augmentation is only performed during the training phase, not in the inference phase.

\begin{equation}\label{eq8}
\begin{aligned}
    \mathcal{L}_{seg} = &-\frac{1}{B\, H\, W}\sum_{j=1}^{B}\sum_{i=1}^{HW} log(p_{ij}) \\
    &+ \frac{1}{BN}\sum_{j=1}^{B} \sum_{n=1}^{N} \left ( 1-\frac{  2 \, \sum_{i=1}^{HW} {p}_{ij}y_{ij}}{ \sum_{i=1}^{HW} ( {p}^{2}_{ij}+y^{2}_{ij} ) } \right )&&
\end{aligned}
\end{equation}
where $H\, W$ is the number of pixels in an image with size $H\times{W}$,  $i=1,2,...,HW$. $B$ is the number of images in a batch, $j=1,2,...,B$. $N$ is the number of classes with $n=1,2,...,N$. $p_{ij}$ and $y_{ij}$ denote the prediction probability of the ground-truth class, and the ground-truth class  of the $i_{th}$ pixel in the $j_{th}$ image, respectively.

\section{Experiments and Analysis}

\label{experiments}
We evaluate the semantic segmentation performance of HSMix on medical images. The datasets and models are presented in Sec. \ref{Data-Model}. Sec. \ref{exp-detail} contains  the implementation details. We then present the results in Sec. \ref{results} and compare the results obtained by training with our HSMiX or without HSMiX for different medical image segmentation tasks and deep neural network models.


\subsection{Models and Datasets for Segmentation}
\label{Data-Model}

The efficacy of our proposed method is evaluated on a range of deep neural network models, including \textcolor{black}{the} UNet \cite{ronneberger2015u}, \textcolor{black}{the} UNet structure with encoder EfficientNet-V2 \cite{tan2019efficientnet}, \textcolor{black}{the} UNet structure with encoder ResNet50 \cite{he2016deep}, \textcolor{black}{the}
UNet structure with encoder ResNet101 \cite{he2016deep}, and
\textcolor{black}{the} DeepLabv3+ \cite{chen2018encoder}. Furthermore, our method is evaluated on \textcolor{black}{the} transformer-based TransUnet  \cite{chen2021transunet} and \textcolor{black}{the} HiFormer \cite{heidari2023hiformer}. In order to align with the evolving landscape of lightweight networks for expedient medical semantic segmentation, our method is also evaluated with \textcolor{black}{the} UNeXt \cite{valanarasu2022unext}.

The evaluation process uses a variety of medical image datasets in different modalities, including the ISIC 2017 T1 dataset \textcolor{black}{with camera images} of skin lesions \cite{codella2018skin}, the GlaS dataset of gland images under the microscope \cite{sirinukunwattana2017gland}, and the MoNuSeg dataset of nuclear images under the microscope \cite{kumar2019multi}. All \textcolor{black}{of} the above datasets are intended for binary segmentation. In addition, the Synapse dataset of CT images is also used for multi-label organ segmentation. \textcolor{black}{The BraTS2018 dataset \footnote{\url{https://www.med.upenn.edu/sbia/BraTS2018/data.html}}, consisting of magnetic resonance imaging (MRI) data, is used for the purpose of multimodal segmentation of brain tumors.}
The ISIC 2017 Task 1 dataset contains 2,000 images for training, 150 images for validation and 600 images for testing. The MoNuSeg dataset contains 30 training images and 14 test images, while the GlaS dataset contains 85 training images and 80 test images.
According to the setting in \cite{chen2021transunet}, the Synapse dataset consists of two subsets: the training dataset with 18 cases and the validation dataset with 12 cases, resulting in a total of 2212 axial slices. \textcolor{black}{The BraTS2018 challenge training data comprises 285 multimodal 3D MRI scans, including 210 MRI scans from subjects with higher grade glioma (HGG) and 75 scans from subjects with lower grade glioma (LGG). The 3D MRI images have a volume dimension of $240\times{240}\times{155}$. They describe four MRI modalities: native (T1), post-contrast T1-weighted (T1Gd), T2-weighted (T2) and T2 Fluid Attenuated Inversion Recovery (FLAIR) volumes. The tumor segmentation consists of dividing the tumor into three distinct sub-regions: the whole tumor (WT), the tumor core (TC) and the enhancing tumor (ET). The dataset was first split into a training and validation set comprising $80\%$ of the data and a test set comprising $20\%$ of the data. Subsequent images were extracted into 2D image slices in accordance with Kermi et al. \cite{kermi2019deep}.}

The metrics for the evaluation are the JAC (mean Jaccard Index), DSC (mean Dice Similarity Coefficient), the HD95 (95th percentile of the Hausdorff Distance).

\subsection{Implementation Details}
\label{exp-detail}
The SLIC algorithm, as described in \cite{achanta2012slic} is employed to generate the superpixel grids from the images. \textcolor{black}{To increase the} augmentation diversity and \textcolor{black}{obtain} more boundaries and contours, we generate superpixels  \textcolor{black}{with different granularity}. In particular, the number of superpixels, denoted as \( l \), is \textcolor{black}{randomly} selected following a uniform distribution. Here, \( l \) follows the distribution \( l \sim U(l_{\text{min}}, l_{\text{max}}) \). The compactness parameter for superpixel generation is set to 0.1 for Synapse and 0.003 for the BraTS2018 dataset, while it is set to 10 for the others datasets. We utilize the Static Saliency Fine Grained function in OpenCV\footnote{\url{https://docs.opencv.org/3.4/da/dd0/classcv_1_1saliency_1_1StaticSaliencyFineGrained.html\#details}} to generate the saliency maps of \textcolor{black}{the} images. In the 2017 Task 1 dataset, the number of superpixels ranges from 30 to 80. The Synapse and GlaS datasets employ a range of 200 to 400 superpixels, while the number of \textcolor{black}{superpixels in the }MoNuSeg dataset  \textcolor{black}{ranges} from 300 to 500.  \textcolor{black}{The} superpixels are selected by a Bernoulli distribution with probability \( p = 0.3 \). Unless otherwise specified, the default input size of \textcolor{black}{the} images and ground truth masks is $224\times{224}$.
Basic data augmentation involves geometric transformations, including randomly flipping or rotating all training images and segmentation masks within an angular range of \( -20^\circ \) to \( 20^\circ \). In the baseline experiments without \textcolor{black}{the} proposed data augmentation,  the ISIC 2017 Task 1 dataset was trained with a batch size of 8 and an initial learning rate of 0.0001. When using the proposed data augmentation, the batch size for the ISIC 2017 Task 1 dataset is set to 4, with an initial learning rate remaining at \(0.0001\). For both experiments, the MoNuSeg dataset was trained with a batch size of 4 and an initial learning rate of \(0.001\). Similarly, the GlaS dataset was trained using a batch size of 4 and an initial learning rate of \(0.0001\). \textcolor{black}{The input size of the 2D image in the BraTS2018 dataset is $224\times{224}$. We perform pre-processing and basic augmentation according to Kermi et al. \cite{kermi2019deep}. The batch size is 4 with an initial learning rate of $0.001$. The number of superpixels ranges from 50 to 150.} 
It is worth noting that the actual batch size is doubled in the experiments with the proposed data augmentation. This is because two augmented images are generated and then used for training. The AdamW optimizer is employed with a weight decay value of $0.0005$.

To obtain reliable results, we run each experiment three times to obtain the results with \textcolor{black}{mean} and standard deviation for ISIC 2017 Task 1. The five-fold cross-validation technique is used for both the MoNuSeg and GlaS datasets.


\subsection{Experimental Results}
\label{results}

\begin{table*}[ht!]
\centering
\caption{\label{Table1} Performance of HSMix with various backbones on the ISIC 2017 Task 1 dataset.}
\setlength{\tabcolsep}{1.5mm}{
\resizebox{0.75\textwidth}{!}{
\begin{tabular}{lccc}
\toprule
\multirow{2}{*}{\textbf{Method}}  & \multicolumn{3}{c}{\textbf{ISIC 2017 T1}}\\ 
\cline{2-4} 
& \textbf{DSC($\%$)$\uparrow$}  & \textbf{JAC($\%$)$\uparrow$} & \textbf{HD95$\downarrow$} \\
    \midrule
    UNet \cite{ronneberger2015u} &  $80.92\pm{0.58}$ & $72.71\pm{0.56}$ & $22.47\pm{0.71}$\\
    UNet + HSMix &  $\mathbf{83.50\pm{0.67}}$ & $\mathbf{74.13\pm{0.51}}$ & $\mathbf{19.82\pm{0.62}}$\\
    \midrule
    UNet-EfficientNet-b2 \cite{tan2019efficientnet} &  $83.52\pm{0.43}$ & $73.77\pm{0.45}$ & $18.51\pm{0.58}$\\
    UNet-EfficientNet-b2 + HSMix &  $\mathbf{84.00\pm{0.29}}$ & $\mathbf{75.16\pm{0.31}}$ & $\mathbf{18.26\pm{0.15}}$\\
    \midrule
    UNet-ResNet50 \cite{he2016deep} &  $83.77\pm{0.30}$ & $74.43\pm{0.41}$ & $19.50\pm{0.53}$\\
    UNet-ResNet50 + HSAMix &  $\mathbf{84.81\pm{0.82}}$ & $\mathbf{76.05\pm{0.78}}$ & $\mathbf{16.37\pm{0.70}}$\\
    \midrule
    UNeXt \cite{valanarasu2022unext} &  $81.14\pm{0.98}$ & $72.71\pm{1.11}$ & $20.16\pm{1.03}$\\
    UNeXt + HSMix &  $\mathbf{82.74\pm{0.37}}$ & $\mathbf{73.91\pm{0.96}}$ & $\mathbf{19.52\pm{0.69}}$\\
    \midrule
    DeepLabv3+ \cite{chen2018encoder} &  $83.33\pm{0.81}$ & $74.27\pm{0.90}$ & $19.21\pm{0.68}$\\
    DeepLabv3+ + HSMix &  $\mathbf{84.69\pm{0.86}}$ & $\mathbf{75.11\pm{0.73}}$ & $\mathbf{16.73\pm{0.74}}$\\
    \midrule
    TransUnet \cite{chen2021transunet} &  $84.76\pm{0.26}$ & $76.52\pm{0.50}$ & $17.38\pm{0.30}$\\
    TransUnet +HSMix &  $\mathbf{85.57\pm{0.46}}$ & $\mathbf{77.34\pm{0.43}}$ & $\mathbf{15.92\pm{0.48}}$\\
    \midrule
    HiFormer-B \cite{heidari2023hiformer} &  $85.43\pm{0.25}$ & $77.14\pm{0.34}$ & $16.88\pm{0.89}$\\
    HiFormer-B + HSMix &  $\mathbf{85.56\pm{0.62}}$ & $\mathbf{77.53\pm{0.59}}$ & $\mathbf{16.54\pm{0.64}}$\\

\bottomrule
\vspace{-1em}
\end{tabular}}}
\end{table*}

Results of "mean$\pm$std" are presented for \textcolor{black}{the} ISIC 2017 T1, \textcolor{black}{the} GlaS and \textcolor{black}{the} MoNuSeg. \textcolor{black}{The} UNet \cite{ronneberger2015u} represents the \textcolor{black}{classic} UNet, while \textcolor{black}{the} UNet-ResNet50 \cite{he2016deep}, \textcolor{black}{the} UNet-EfficientNet-b2 \cite{tan2019efficientnet} \textcolor{black}{and the} UNet-ResNet101 \cite{he2016deep} represent the architecture of \textcolor{black}{the} UNet with altered encoders.

\begin{table*}[ht!]
\centering
\caption{\label{Table2} Performance of HSMix with various backbones on the GlaS dataset.}
\setlength{\tabcolsep}{1.5mm}{
\resizebox{0.8\textwidth}{!}{
\begin{tabular}{lccc}
\toprule
\multirow{2}{*}{\textbf{Method}}  & \multicolumn{3}{c}{\textbf{GlaS}}\\ 
\cline{2-4} 
& \textbf{DSC($\%$)$\uparrow$}  & \textbf{JAC($\%$)$\uparrow$} & \textbf{HD95$\downarrow$} \\
    \midrule
    UNet \cite{ronneberger2015u} &  $85.10\pm{0.69}$ & $75.22\pm{1.17}$ & $26.30\pm{0.32}$\\
    UNet + HSMix &  $\mathbf{90.83\pm{0.90}}$ & $\mathbf{83.97\pm{1.36}}$ & $\mathbf{15.61\pm{1.15}}$\\
    \midrule
    UNet-EfficientNet-b2 \cite{tan2019efficientnet} &  $87.93\pm{1.64}$ & $79.39\pm{2.61}$ & $20.52\pm{3.30}$\\
    UNet-EfficientNet-b2 + HSMix &  $\mathbf{88.85\pm{0.93}}$ & $\mathbf{80.15\pm{1.33}}$ & $\mathbf{20.46\pm{1.90}}$\\
    \midrule
    UNet-ResNet50 \cite{he2016deep} &  $90.28\pm{1.12}$ & $83.53\pm{1.55}$ & $16.43\pm{0.98}$\\
    UNet-ResNet50 + HSMix &  $\mathbf{90.51\pm{0.47}}$ & $\mathbf{83.74\pm{0.94}}$ & $\mathbf{16.41\pm{0.61}}$\\
    \midrule
    UNeXt \cite{valanarasu2022unext} &  $82.00\pm{1.86}$ & $70.96\pm{2.44}$ & $58.89\pm{4.65}$\\
    UNeXt + HSMix &  $\mathbf{86.20\pm{0.88}}$ & $\mathbf{76.85\pm{1.23}}$ & $\mathbf{21.04\pm{0.68}}$\\
    \midrule
    DeepLabv3+ \cite{chen2018encoder} &  $90.09\pm{1.10}$ & $82.79\pm{1.26}$ & $15.70\pm{0.63}$\\
    DeepLabv3+ + HSMix &  $\mathbf{90.28\pm{0.79}}$ & $\mathbf{82.95\pm{1.17}}$ & $\mathbf{15.45\pm{0.56}}$\\
    \midrule
    TransUnet \cite{chen2021transunet} &  $89.79\pm{0.82}$ & $82.88\pm{1.46}$ & $15.05\pm{1.49}$\\
    TransUnet + HSMix &  $\mathbf{90.95\pm{0.35}}$ & $\mathbf{84.18\pm{0.51}}$ & $\mathbf{14.71\pm{0.45}}$\\
    \midrule
    HiFormer-B \cite{heidari2023hiformer} &  $90.55\pm{0.48}$ & $83.47\pm{0.76}$ & $14.79\pm{0.88}$\\
    HiFormer-B + HSMix &  $\mathbf{90.92\pm{0.28}}$ & $\mathbf{84.03\pm{0.43}}$ & $\mathbf{14.31\pm{0.92}}$\\

\bottomrule
\vspace{-1em}
\end{tabular}}}
\end{table*}

Tables
\ref{Table1}, \ref{Table2}, \ref{Table3}, \ref{Table4}  and \ref{Table4-add} illustrate the enhanced performance of HSMix across diverse segmentation tasks and models. \textcolor{black}{The enhancement is the outcome of a combination of hard and soft mixing for data augmentation. Our proposed HSMix not only increases the diversity of the training space, but also facilitates the smoothing of the labeling space by incorporating rich contour and saliency information.} The effectiveness of HSMix has been demonstrated on binary medical segmentation tasks (see Tables
\ref{Table1}, \ref{Table2}, \ref{Table3}) as well as on medical segmentation tasks with multiple labels(see Table \ref{Table4}, Table \ref{Table4-add}). It has been shown that our HSMix method improves performance of both CNN-based and Transformer-based models. In particular, Table \ref{Table1} indicates that HSMix increases the DSC from  $80.92\%$ to $83.50\%$ with \textcolor{black}{the} UNet and from $84.76\%$ to $85.57\%$ with \textcolor{black}{the} TransUnet for the ISIC 2017 T1 dataset. For the GlaS dataset, Table \ref{Table2} shows that the JAC increases from  $79.39\%$ to $80.15\%$ with \textcolor{black}{the}  UNet-EfficientNet-b2 and increases $83.47\%$ to $84.03\%$ with \textcolor{black}{the}  HiFormer-B model. Additionally, our HSMix is effective for both lightweight and fast models such as \textcolor{black}{the} UNeXt and \textcolor{black}{the} classical models like DeepLabv3+. In particular, Table \ref{Table3} indicates that the DSC increases from $71.61\%$ to $72.29\%$, the JAC increases from $56.24\%$ to $56.84\%$ with \textcolor{black}{the} UNeXt network. With DeepLabv3+, it can be observed in Table \ref{Table3} that JAC increases from $49.94\%$ to $50.28\%$ and the HD95 decreases from $3.66$ to $3.55$. For the multi-label Synapse dataset, Table \ref{Table4} shows that our HSMix increases the DSC from $77.59\%$ to $78.25\%$ and the JAC from $64.24\%$ to $68.40\%$ with \textcolor{black}{the} UNet network. With \textcolor{black}{the} TransUnet, the DSC increases from $77.74\%$ to $79.35\%$ and the JAC from $67.46\%$ to $69.30\%$. \textcolor{black}{Table \ref{Table4-add} illustrates that our proposed data augmentation method remains effective even when applied to more challenging MRI data. Additionally, Table \ref{Table4-add}  provides an evaluation of the computational overhead associated with our proposed method. As shown in Table \ref{Table4-add}, the implementation of our proposed HSMix only lead to an increase in training time. However, the size of the model (Param.) and the computational complexity of the deep model (FLOPs) do not increase. The proposed data augmentation method, HSMix, is only applied during the training phase and not during the inference phase. Consequently, the application of HSMix has no impact on the inference time (FPS).}

\begin{table*}[t]
\centering
\caption{\label{Table3} Performance of HSMix with various backbones on the MoNuSeg dataset.}
\setlength{\tabcolsep}{1.5mm}{
\resizebox{0.8\textwidth}{!}{
\begin{tabular}{lccc}
\toprule
\multirow{2}{*}{\textbf{Method}}  & \multicolumn{3}{c}{\textbf{MoNuSeg}}\\ 
\cline{2-4} 
& \textbf{DSC($\%$)$\uparrow$}  & \textbf{JAC($\%$)$\uparrow$} & \textbf{HD95$\downarrow$} \\
    \midrule
    UNet \cite{ronneberger2015u} &  $76.51\pm{2.88}$ & $63.12\pm{2.34}$ & $3.69\pm{1.17}$\\
    UNet + HSMix &  $\mathbf{79.63\pm{0.73}}$ & $\mathbf{66.68\pm{0.69}}$ & $\mathbf{2.77\pm{0.08}}$\\
    \midrule
    UNet-EfficientNet-b2 \cite{tan2019efficientnet} &  $76.08\pm{1.22}$ & $61.58\pm{1.53}$ & $4.34\pm{0.35}$\\
    UNet-EfficientNet-b2 + HSMix &  $\mathbf{78.63\pm{0.89}}$ & $\mathbf{65.21\pm{1.03}}$ & $\mathbf{3.21\pm{0.22}}$\\
    \midrule
    UNet-ResNet50 \cite{he2016deep} &  $77.37\pm{1.01}$ & $63.30\pm{1.34}$ & $3.90\pm{0.41}$\\
    UNet-ResNet50 + HSMix &  $\mathbf{77.67\pm{1.24}}$ & $\mathbf{64.51\pm{1.45}}$ & $\mathbf{3.15\pm{0.26}}$\\
    \midrule
    UNeXt \cite{valanarasu2022unext} &  $71.61\pm{1.31}$ & $56.24\pm{1.49}$ & $11.72\pm{0.40}$\\
    UNeXt + HSMix &  $\mathbf{72.29\pm{2.11}}$ & $\mathbf{56.84\pm{2.43}}$ & $\mathbf{11.18\pm{0.75}}$\\
    \midrule
    DeepLabv3+ \cite{chen2018encoder} &  $66.28\pm{0.55}$ & $49.94\pm{0.56}$ & $3.66\pm{0.34}$\\
    DeepLabv3+ HSMix &  $\mathbf{66.53\pm{0.89}}$ & $\mathbf{50.28\pm{0.97}}$ & $\mathbf{3.55\pm{0.52}}$\\
    \midrule
    TransUnet \cite{chen2021transunet} &  $78.55\pm{0.93}$ & $65.06\pm{0.73}$ & $2.91\pm{0.72}$\\
    TransUnet + HSMix &  $\mathbf{79.63\pm{0.82}}$ & $\mathbf{66.28\pm{1.11}}$ & $\mathbf{2.50\pm{0.21}}$\\
    \midrule
    HiFormer-B \cite{heidari2023hiformer} &  $68.06\pm{0.86}$ & $51.89\pm{1.11}$ & $3.15\pm{0.27}$\\
    HiFormer-B + HSMix &  $\mathbf{68.81\pm{1.26}}$ & $\mathbf{52.71\pm{0.97}}$ & $\mathbf{3.10\pm{0.58}}$\\

\bottomrule
\vspace{-1em}
\end{tabular}}}
\end{table*}

\begin{table*}[h]
\centering
\caption{\label{Table4} Performance of HSMix with various backbones on the Synapse dataset.}
\setlength{\tabcolsep}{1.5mm}{
\resizebox{1.0\textwidth}{!}{
\begin{tabular}{lccc|cccccccc}
\toprule
\multirow{2}{*}{\textbf{Method}}  & \multicolumn{3}{c}{\textbf{Average Performance}} & \multicolumn{8}{c}{\textbf{DSC ($\%$) for each class}}\\ 
\cline{2-12} 
& \textbf{DSC ($\%$)$\uparrow$}  & \textbf{JAC ($\%$)$\uparrow$} & \textbf{HD95 $\downarrow$} &Aorta&Gallbladder&Kidney(L)&Kidney(R)&Liver&Pancreas&Spleen&Stomach\\
    \midrule
    UNet \cite{ronneberger2015u}  &  $77.59$ & $64.24$ & $34.45$ & $85.89$ & $60.25$ & $81.74$ & $78.03$ & $94.80$ & $60.90$ & $87.89$ & $71.21$\\
    UNet + HSMix  &  $\mathbf{78.25}$ & $\mathbf{68.40}$ & $\mathbf{29.51}$ & $86.16$ & $64.75$ & $82.47$ & $79.41$ & $95.05$ & $59.71$ & $85.59$ & $72.85$\\
    \midrule
    UNet-EfficientNet-b2 \cite{tan2019efficientnet}  &  $74.58$ & $64.45$ & $33.59$ & $86.86$ & $60.75$ & $80.82$ & $68.54$ & $93.98$ & $46.91$ & $86.09$ & $72.72$\\
    UNet-EfficientNet-b2 + HSMix &  $\mathbf{75.26}$ & $\mathbf{64.30}$ & $\mathbf{28.91}$ & $82.71$ & $60.39$ & $79.98$ & $73.25$ & $94.26$ & $59.04$ & $85.63$ & $66.77$\\
    \midrule
    UNet-ResNet50 \cite{he2016deep} &  $70.89$ & $60.53$ & $45.96$ & $85.49$ & $53.54$ & $74.11$ & $67.03$ & $93.43$ & $41.90$ & $86.32$ & $65.29$\\
    UNet-ResNet50 + HSMix  &  $\mathbf{73.13}$ & $\mathbf{62.44}$ & $\mathbf{44.08}$ & $75.52$ & $55.40$ & $78.46$ & $69.98$ & $94.19$ & $54.22$ & $79.24$ & $67.57$\\
    \midrule
    UNeXt \cite{valanarasu2022unext}  &  $65.43$ & $54.10$ & $33.41$ & $71.56$ & $47.00$ & $70.70$ & $66.95$ & $92.52$ & $31.52$ & $80.25$ & $62.97$\\
    UNeXt + HSMix  &  $\mathbf{66.22}$ & $\mathbf{54.71}$ & $\mathbf{31.44}$ & $78.30$ & $47.80$ & $67.54$ & $64.15$ & $92.29$ & $35.46$ & $76.15$ & $68.02$\\
    \midrule
    DeepLabv3+ \cite{chen2018encoder}  &  $70.58$ & $60.10$ & $31.17$ & $77.90$ & $52.02$ & $78.52$ & $70.85$ & $93.28$ & $38.56$ & $86.60$ & $66.90$\\
    DeepLabv3+ + HSMix  &  $\mathbf{70.78}$ & $\mathbf{60.67}$ & $\mathbf{30.91}$ & $79.33$ & $43.49$ & $80.01$ & $74.17$ & $93.85$ & $36.39$ & $87.16$ & $71.88$\\
    \midrule
    TransUnet \cite{chen2021transunet}  &  $77.74$ & $67.46$ & $33.76$ & $87.89$ & $64.64$ & $83.27$ & $78.05$ & $94.86$ & $52.57$ & $86.64$ & $74.08$\\
    TransUnet + HSMix  &  $\mathbf{79.35}$ & $\mathbf{69.30}$ & $\mathbf{28.76}$ & $86.82$ & $61.76$ & $81.57$ & $76.52$ & $94.94$ & $66.67$ & $88.22$ & $78.31$\\
    \midrule
    HiFormer-B \cite{heidari2023hiformer}  &  $70.23$ & $59.18$ & $34.76$ & $75.56$ & $52.31$ & $76.93$ & $73.74$ & $92.90$ & $34.19$ & $86.10$ & $70.10$\\
    HiFormer-B + HSMix &  $\mathbf{71.53}$ & $\mathbf{60.50}$ & $\mathbf{34.12}$ & $76.73$ & $50.64$ & $74.07$ & $75.04$ & $93.39$ & $45.91$ & $87.27$ & $69.17$\\
\bottomrule
\vspace{-1em}
\end{tabular}}}
\end{table*}

\begin{table*}[ht!]
\centering
\caption{\label{Table4-add} \textcolor{black}{
Performance of HSMix with different backbones on the BraTS2018 dataset. "WT, TC, ET" are the three tumor categories. "Param." indicates the model size, "FLOPs" denotes the computational complexity per image with deep model, "Training Time" represents the  time in seconds  required to train one epoch, "FPS" is the number of frames per second, which indicates the inference speed. The evaluation is performed with Intel Core i9-13900KF and NVIDIA GeForce RTX 4090.}}

\setlength{\tabcolsep}{1.5mm}{
\resizebox{1\textwidth}{!}{
\begin{tabular}{lcccc|cccc|rrrr}
\toprule
\multirow{2}{*}{\textbf{Method}}  & \multicolumn{4}{c}{\textbf{DSC ($\%$)$\uparrow$}} & \multicolumn{4}{c}{\textbf{JAC ($\%$)$\uparrow$}} & \multicolumn{4}{c}{\textbf{Computational Overhead}}\\ 
\cline{2-13} 
& \textbf{Avg.}  & \textbf{WT} & \textbf{TC} & \textbf{ET} 
& \textbf{Avg.}  & \textbf{WT} & \textbf{TC} & \textbf{ET} 
& \textbf{Param. (M)}$\downarrow$ & \textbf{FLOPs (G)}$\downarrow$ & \textbf{Training Time (s)$\downarrow$} & \textbf{FPS}$\uparrow$\\
    \midrule
    UNet   &  $77.33$ & $83.10$ & $78.30$ & $70.59$ & $69.96$ & $76.38$ & $73.29$ & $60.21$ & $34.52$ & $50.20$ & $196$ & $117$\\
    UNet + HSMix  &  $\mathbf{78.97}$ & $\mathbf{84.36}$ & $\mathbf{80.22}$ & $\mathbf{72.35}$ & $\mathbf{71.35}$ & $\mathbf{77.58}$ & $\mathbf{74.96}$ & $\mathbf{62.07}$ & $34.52$ & $50.20$ & $935$ & $117$\\
    \midrule
    UNeXt  &  $66.41$ & $75.10$ & $65.67$ & $58.45$ & $57.49$ & $66.67$ & $59.14$ & $46.65$ & $1.47$ & $0.44$ & $35$ & $679$\\
    UNeXt + HSMix  &  $\mathbf{68.29}$ & $\mathbf{76.04}$ & $\mathbf{68.27}$ & $\mathbf{60.57}$ & $\mathbf{59.33}$ & $\mathbf{67.44}$ & $\mathbf{61.70}$ & $\mathbf{48.85}$ & $1.47$ & $0.44$ & $642$ & $679$\\

\bottomrule
\vspace{-1em}
\end{tabular}}}
\end{table*}

\begin{table*}[h]
\centering
\caption{\label{Table5} \textcolor{black}{Comparison of model size (Param.), computational complexity (FLOPs) per image, the required  time in  seconds for training one epoch (Training Time), inference speed of frames per second (FPS),  and performance for segmentation on the ISIC 2017 T1 dataset with UNet-ResNet101. The results marked with $\dagger$ are derived from the original publication. The optimal and second-best results are highlighted in $\mathbf{bold}$ and \underline{underlined}, respectively. 
The evaluation is performed with Intel Core i9-13900KF and NVIDIA GeForce RTX 4090.}}
\setlength{\tabcolsep}{1.5mm}{
\resizebox{1.0\textwidth}{!}{
\begin{tabular}{lcccccc}
\toprule
\multirow{2}{*}{\textbf{Method}}  & \multicolumn{4}{c}{\textcolor{black}{\textbf{Computational Overhead}}} & \multicolumn{2}{c}{\textbf{Performance}}\\ 
\cline{2-7} 
& \textcolor{black}{\textbf{Param. (M)}$\downarrow$} & \textcolor{black}{\textbf{FLOPs (G)}$\downarrow$} & \textcolor{black}{\textbf{Training Time (s)$\downarrow$}} & \textcolor{black}{\textbf{FPS}$\uparrow$} & \textbf{DSC ($\%$)$\uparrow$}  & \textbf{JAC ($\%$)$\uparrow$}  \\
    \midrule
    Baseline  & \textcolor{black}{$166.8007$} & \textcolor{black}{$44.85$} & \textcolor{black}{$153$} & \textcolor{black}{$81$} &  $84.38\pm{0.91}$ & $75.45\pm{0.85}$ \\
    MixUp \cite{zhang2018mixup} (Augment Image and Mask) & \textcolor{black}{$166.8007$} & \textcolor{black}{$44.85$} & \textcolor{black}{$156$} & \textcolor{black}{$81$} & $85.40\pm{0.50}$ & $77.04\pm{0.41}$ \\
    MixUp \cite{zhang2018mixup} (Augment Only Image) & \textcolor{black}{$166.8007$} & \textcolor{black}{$44.85$} & \textcolor{black}{$154$} & \textcolor{black}{$81$} &   $85.26\pm{0.56}$ & $76.93\pm{0.37}$ \\
    GAN-model-based \cite{bi2019improving} & \textcolor{black}{$166.8008$} & \textcolor{black}{$89.71$} & \textcolor{black}{$307$} & \textcolor{black}{$81$} &   $85.16\dagger$ & $77.14\dagger$ \\
    CutMix \cite{yun2019cutmix} & \textcolor{black}{$166.8007$} & \textcolor{black}{$44.85$} & \textcolor{black}{$158$} & \textcolor{black}{$81$} &  $85.04\pm{0.69}$ & $76.55\pm{0.51}$\\
    SPDA \cite{zhang2019spda} & \textcolor{black}{$166.8007$} & \textcolor{black}{$44.85$} & \textcolor{black}{$191$} & \textcolor{black}{$81$} &  $54.47\pm{0.68}$ & $39.93\pm{0.83}$ \\
    CutOut \cite{devries2017improved} & \textcolor{black}{$166.8007$} & \textcolor{black}{$44.85$} & \textcolor{black}{$155$} & \textcolor{black}{$81$} &  $84.68\pm{0.57}$ & $76.15\pm{0.52}$\\
    \textcolor{black}{LCAMix \cite{sun2024lcamix}} & \textcolor{black}{$166.8009$} & \textcolor{black}{$44.86$} & \textcolor{black}{$266$} & \textcolor{black}{$81$} &  \textcolor{black}{$\underline{85.83\pm{0.61}}$}  & \textcolor{black}{$\underline{77.35\pm{0.55}}$}\\
    \textbf{Ours} & \textcolor{black}{$166.8007$} & \textcolor{black}{$44.85$} & \textcolor{black}{$262$} & \textcolor{black}{$81$} &  $\mathbf{85.91\pm{0.62}}$ & $\mathbf{77.48\pm{0.73}}$ \\

\bottomrule
\vspace{-1em}
\end{tabular}}}
\end{table*}

\textcolor{black}{Table \ref{Table5} shows a comparative analysis of our HSMix with other augmentation approaches. We evaluate both the performance efficiency and computational cost of HSMix compared to other representative augmentation methods. As shown in Table \ref{Table5}, the Mixup method performs better on the segmentation task with both augmented images and the corresponding augmented labels than the Mixup method with augmented images only. It is important to emphasize that the comparison methods for augmentation, namely Mixup, CutOut and CutMix, are typically used in the context of image classification tasks. In this study, we apply the above augmentation methods to perform medical segmentation tasks and present the results. Moreover, our method outperforms the GAN model-based augmentation method proposed in \cite{bi2019improving}, which is complex and time-consuming and lead to a large increase in FLOPs and training time. SPDA \cite{zhang2019spda} performs significantly worse than the baseline. This can be explained by the fact that although the superpixelization preserves the oversegmented contours, the detailed information for each superpixel is also lost. Compared to LCAMix \cite{sun2024data}, our HSMix demonstrates the ability to improve performance while reducing computational effort.}

\section{Ablation Study}
\label{ablation}
We first examine the effects of hard and soft mixing augmentation, as described separately in Sec \ref{hard_soft}. Subsequently, the number and selection of superpixels are investigated in Sec \ref{nb_selection}. In addition, the effects of the size of the input image are examined in Sec \ref{input_size}. Finally, we present a visual comparison of the predicted segmentation masks in Sec \ref{comparisions}. The results of the ablation study are summarized in Table \ref{Table6}.

\begin{table}[ht!]
\centering
\caption{\label{Table6} Ablation study on the MoNuSeg and GlaS datasets with UNet. Here, the term "Square" indicates a $7\times7$ \textcolor{black}{square} grid, while "Superpixel" refers to the superpixel grid in our method. The term "Random" signifies a randomly selected mixing ratio for the entire image, as the original Mixup method does. The term "Saliency" denotes the mixing ratios that originate from the saliency-based relationship between the pairwise images, as Guided Mixup \cite{kang2023guidedmixup} does.}

\setlength{\tabcolsep}{1.5mm}
\resizebox{1.0\textwidth}{!}{
\begin{tabular}{lc|cccc|rrrr}
\toprule
\multirow{2}{*}{\textbf{NO.}} &
\multirow{2}{*}{\textbf{Mixing Method}} &
\multirow{2}{*}{\textbf{Square}} &
\multirow{2}{*}{\textbf{Superpixel}} &
\multirow{2}{*}{\textbf{Random}} &
\multirow{2}{*}{\textbf{Saliency}} &
\multicolumn{2}{c}{\textbf{GlaS}} &
\multicolumn{2}{c}{\textbf{MoNuSeg}} \\
\cline{7-10}
& & & & & &
\textbf{DSC($\%$)$\uparrow$} & \textbf{JAC($\%$)$\uparrow$} &
\textbf{DSC($\%$)$\uparrow$} & \textbf{JAC($\%$)$\uparrow$} \\
\midrule
1 & Baseline & \ding{56} & \ding{56} & \ding{56} & \ding{56} &
$85.10\pm{0.69}$ & $75.22\pm{1.17}$ & $76.51\pm{2.88}$ & $63.12\pm{2.34}$ \\
\midrule
2 & \multirow{2}{*}{Hard Mixing} & \ding{52} & \ding{56} & \ding{56} & \ding{56} &
$87.17\pm{1.25}$ & $78.83\pm{1.05}$ & $76.83\pm{4.58}$ & $63.37\pm{1.99}$ \\
3 & & \ding{56} & \ding{52} & \ding{56} & \ding{56} &
$88.53\pm{0.64}$ & $80.71\pm{0.84}$ & $78.02\pm{1.35}$ & $64.62\pm{1.51}$ \\
\midrule
4 & \multirow{2}{*}{Soft Mixing} & \ding{56} & \ding{56} & \ding{52} & \ding{56} &
$89.66\pm{0.63}$ & $81.97\pm{0.98}$ & $78.58\pm{1.36}$ & $65.19\pm{1.54}$ \\
5 & & \ding{56} & \ding{56} & \ding{56} & \ding{52} &
$89.72\pm{0.81}$ & $81.99\pm{0.86}$ & $78.69\pm{1.29}$ & $65.26\pm{1.61}$ \\
\midrule
6 & \textbf{Ours} & \ding{56} & \ding{52} & \ding{56} & \ding{52} &
$\mathbf{90.83\pm{0.90}}$ & $\mathbf{83.97\pm{1.36}}$ &
$\mathbf{79.63\pm{0.73}}$ & $\mathbf{66.68\pm{0.69}}$ \\
\bottomrule
\end{tabular}
}
\end{table}

 \subsection{Hard and Soft Mixing Data Augmentation}
\label{hard_soft}

Table \ref{Table6} shows the results of the ablation study, in which the improvement in performance is analyzed.  As shown in Table \ref{Table6}, both hard and soft mixing can improve performance compared to the baseline, when used separately. \textcolor{black}{This is due to the fact that both hard and soft mixing can enhance the diversity of the training datasets. As can be seen in Table \ref{Table6}, single soft pixel-wise mixing performs better than the single hard mixing with smooth labeling in the augmented mask.}  Hard mixing with superpixel grids (Row 3 compared to Row 1) results in an increase in DSC from $85.10\%$ to $88.53\%$ and an increase in JAC from $75.22\%$ to $80.71\%$ for the GlaS dataset compared to the baseline. Soft mixing with saliency (Row 5 compared to Row 1) led to an increase in DSC from $85.10\%$ to $89.72\%$ and an increase in JAC from $75.22\%$ to $81.99\%$ for the GlaS dataset compared to the baseline. \textcolor{black}{It can also be observed that the use of superpixel grids instead of square grids results in the effect of generating an augmented image, which can improve performance (Row 3 compared to Row 2).} \textcolor{black}{Furthermore, the saliency-based mixup (Row 5) demonstrates an improvement in the performance of the traditional soft mixup method (Row 4) by including more saliency information. In the traditional soft mixup method (Row 4) a value between 0 to 1 is randomly selected as the mixing ratio for the entire image.} In contrast, in saliency-based mixup (Row 5) the mixing ratio for each superpixel is created by considering the relative saliency map of the pairwise images. We emphasize that the soft mixing shown in Rows 4 and 5 does not use any superpixel grid. Instead, the mixing coefficients are global (Row 4) or pixel-dependent (Row 5).  

The proposed method (Row 6) involves both hard and soft mixing, which can complement each other. \textcolor{black}{Hard mixing provides exact labeling (binary labeling), while soft mixing provides smooth labeling (continuous value) in the augmented mask. The combination of these two techniques not only expands the image space but also smooths the mask space, allowing for a more diverse and robust augmentation.} As shown in Table \ref{Table6},  the combination of hard and soft mixing (Row 6) significantly improved enhanced the performance compared to the baseline (Row 1), with the DSC increasing from $76.51\%$ to $79.63\%$ and the JAC from $63.12\%$ to $66.68\%$ for \textcolor{black}{the} MoNuSeg dataset.

\subsection{Generation and Selection of Superpixels}
\label{nb_selection}

\textcolor{black}{The creation and selection of the superpixel map depends on three important parameters. Two parameters are implicitly related to the superpixel segmentation algorithm SLIC \cite{achanta2012slic}: (i) the number of superpixels, $l$, which determines the granularity of the local regions, and (ii)  the compactness parameter, $c$, which balances the color proximity and spatial proximity in the SLIC algorithm. The third parameter is the selection probability of superpixels, $p$, which influences the number of superpixels selected for blending. All these factors are decisive for the mixing process.}

\begin{figure*}[ht!]
      \centering
  \begin{subfigure}[H]{0.19\linewidth}
		\centering
  \includegraphics[height=2.3cm,width=2.3cm]{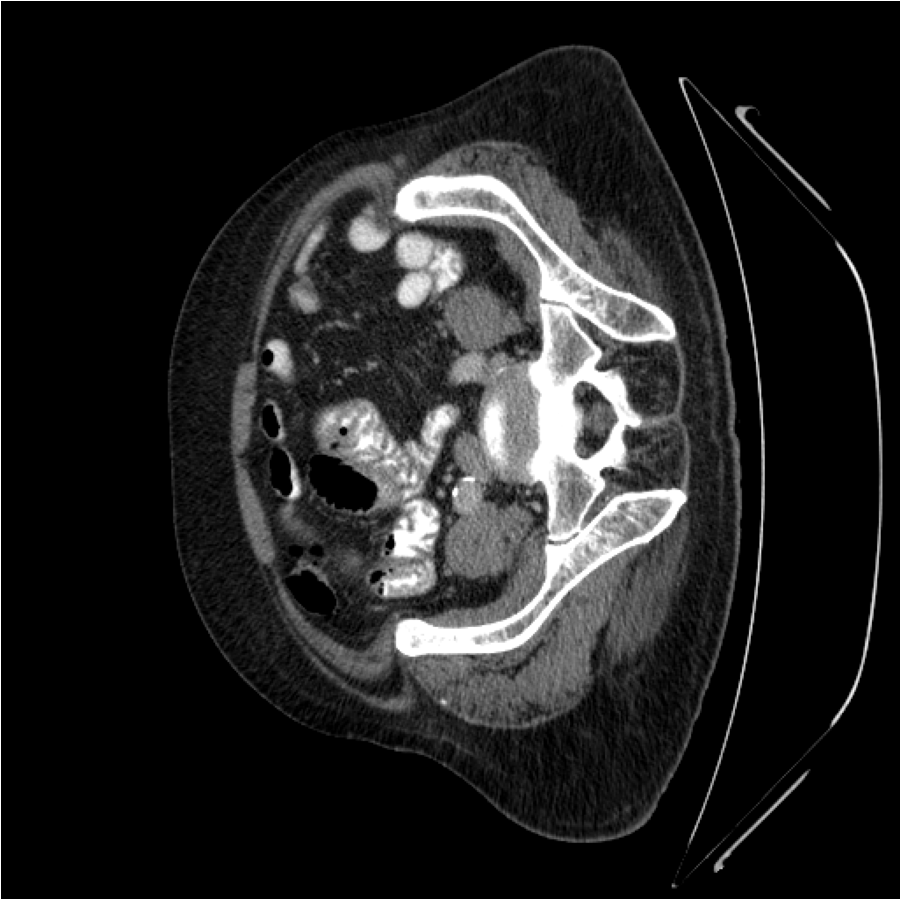}
		\caption{Img. $\Xvect_1$}
        \label{fig4:a} 
	\end{subfigure} 
 \begin{subfigure}[H]{0.19\linewidth}
		\centering
  \includegraphics[height=2.3cm,width=2.3cm]{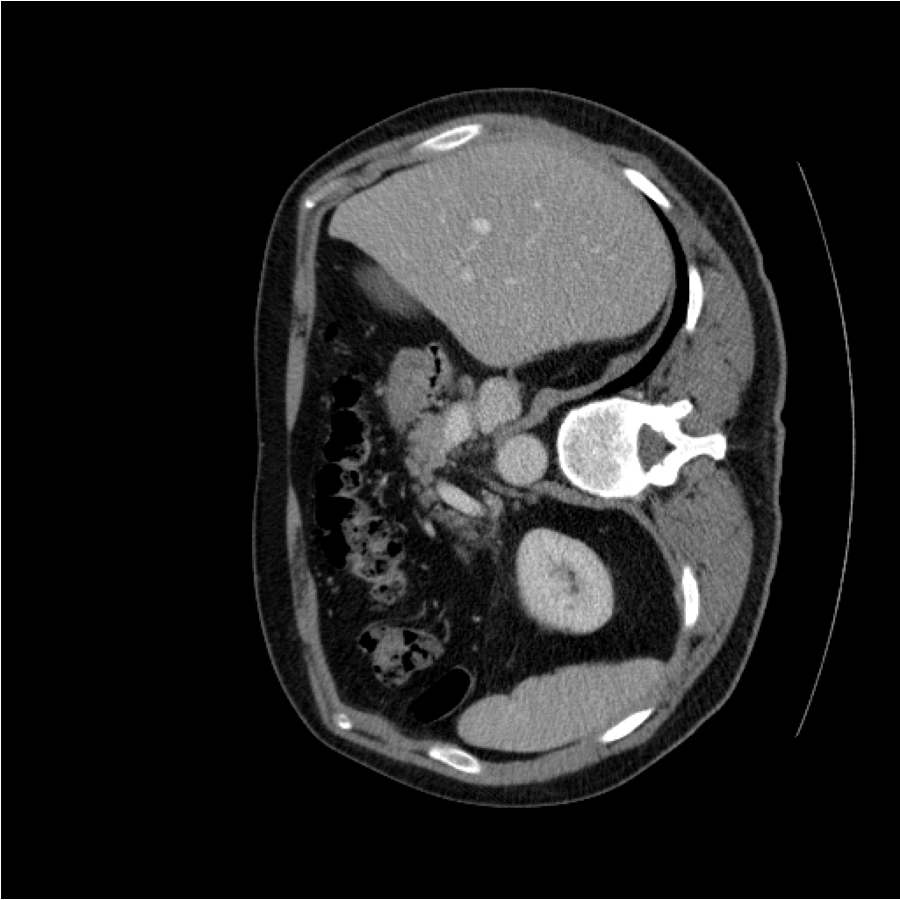}
		\caption{Img. $\Xvect_2$}
        \label{fig4:b} 
	\end{subfigure}
\begin{subfigure}[H]{0.19\linewidth}
		\centering
  \includegraphics[height=2.3cm,width=2.3cm]{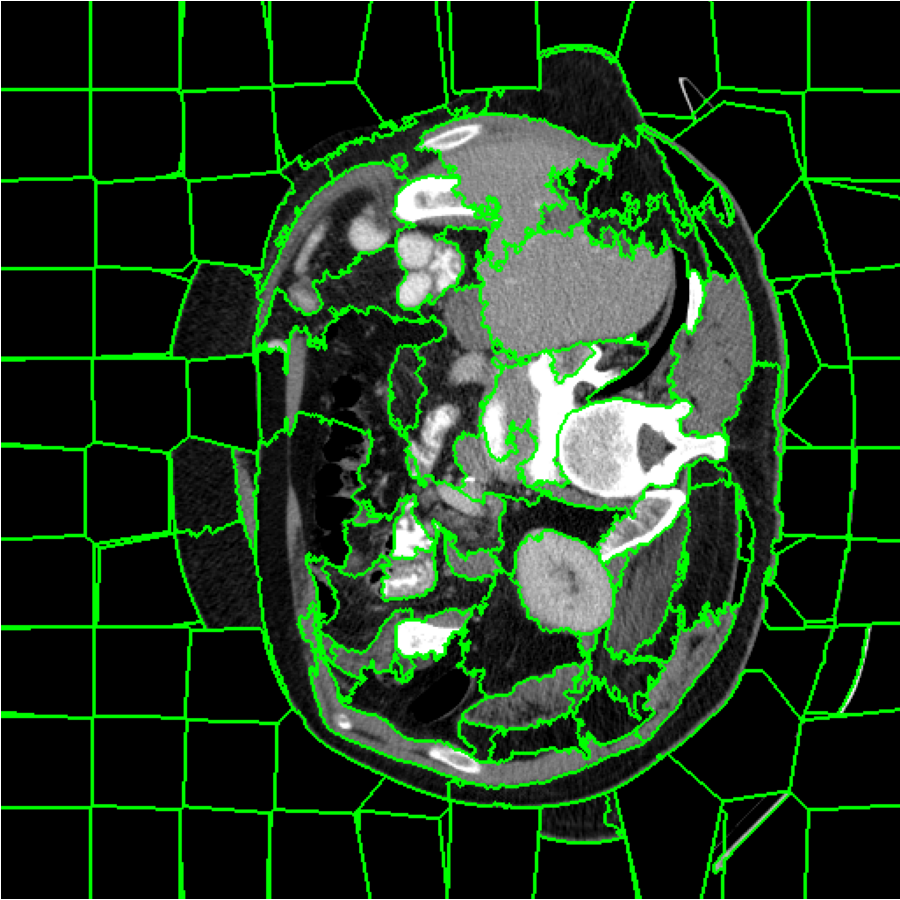}
		\caption{$l_{1}=l_{2}=100$}
        \label{fig4:c} 
	\end{subfigure}
\begin{subfigure}[H]{0.19\linewidth}
		\centering
  \includegraphics[height=2.3cm,width=2.3cm]{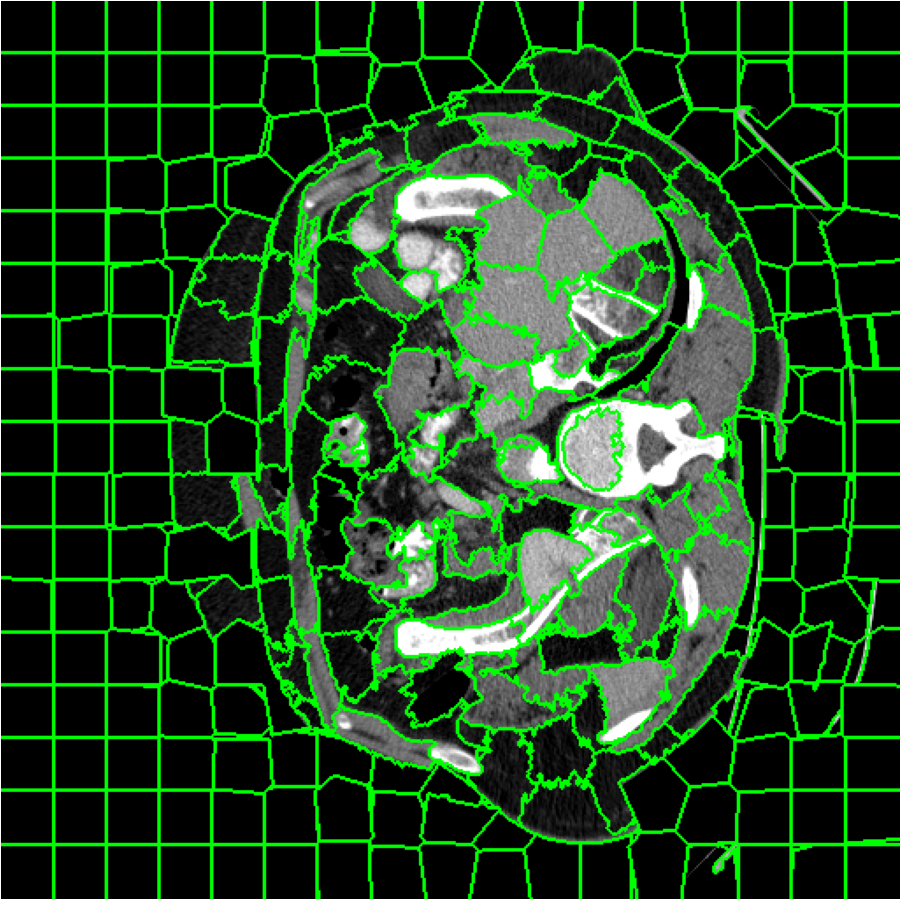}
		\caption{$l_{1}=l_{2}=300$}
        \label{fig4:d} 
	\end{subfigure}
\begin{subfigure}[H]{0.19\linewidth}
		\centering
 \includegraphics[height=2.3cm,width=2.3cm]{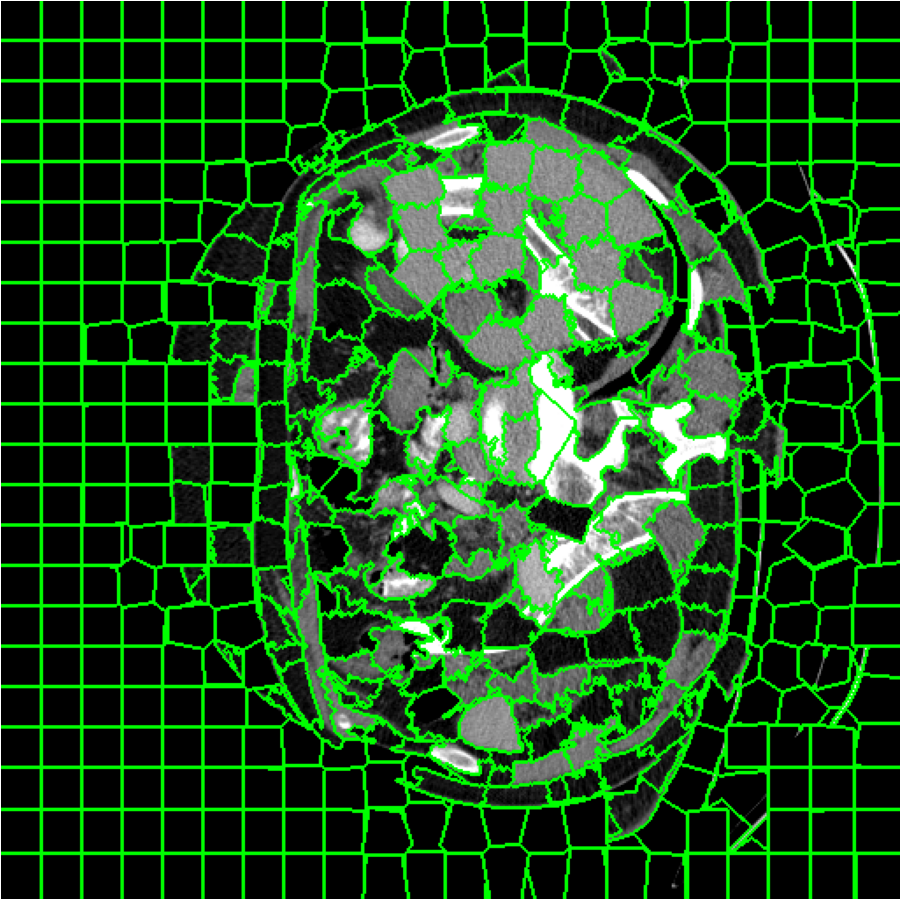}
		\caption{$l_{1}=l_{2}=500$}
        \label{fig4:e} 
	\end{subfigure}

	\begin{subfigure}[H]{0.19\linewidth}
		\centering
  \includegraphics[height=2.3cm,width=2.3cm]{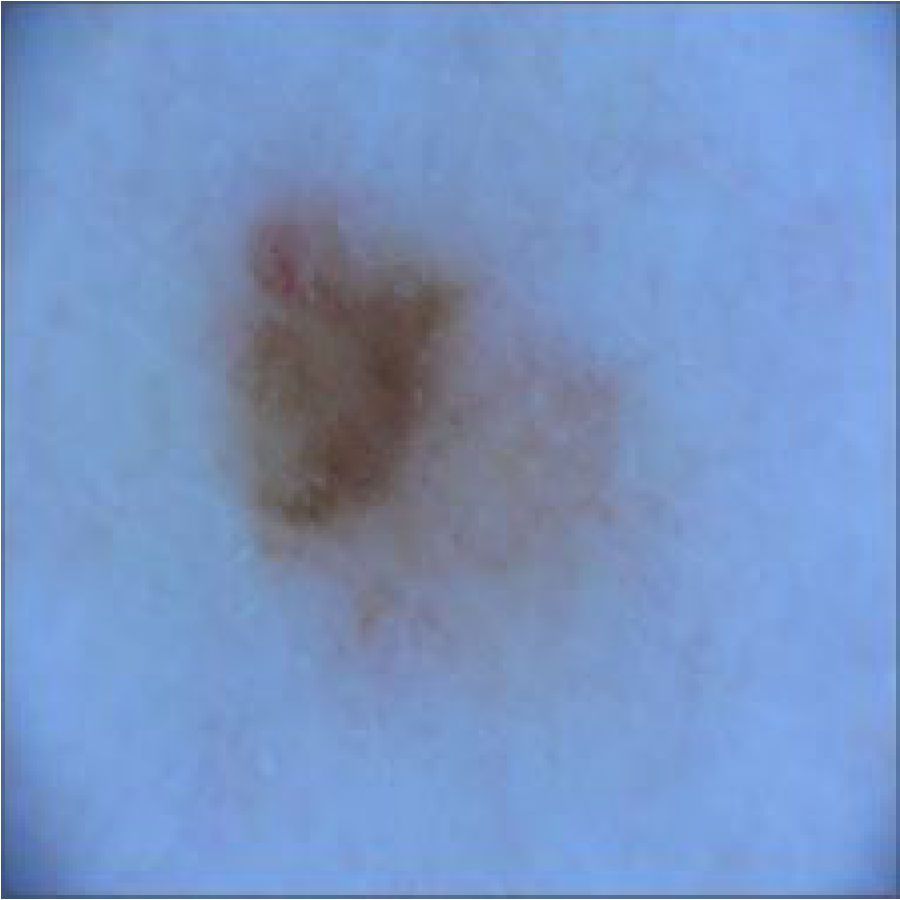}
		\caption{Img. $\Xvect_1$}
        \label{fig4:a1} 
	\end{subfigure} 
 \begin{subfigure}[H]{0.19\linewidth}
		\centering
  \includegraphics[height=2.3cm,width=2.3cm]{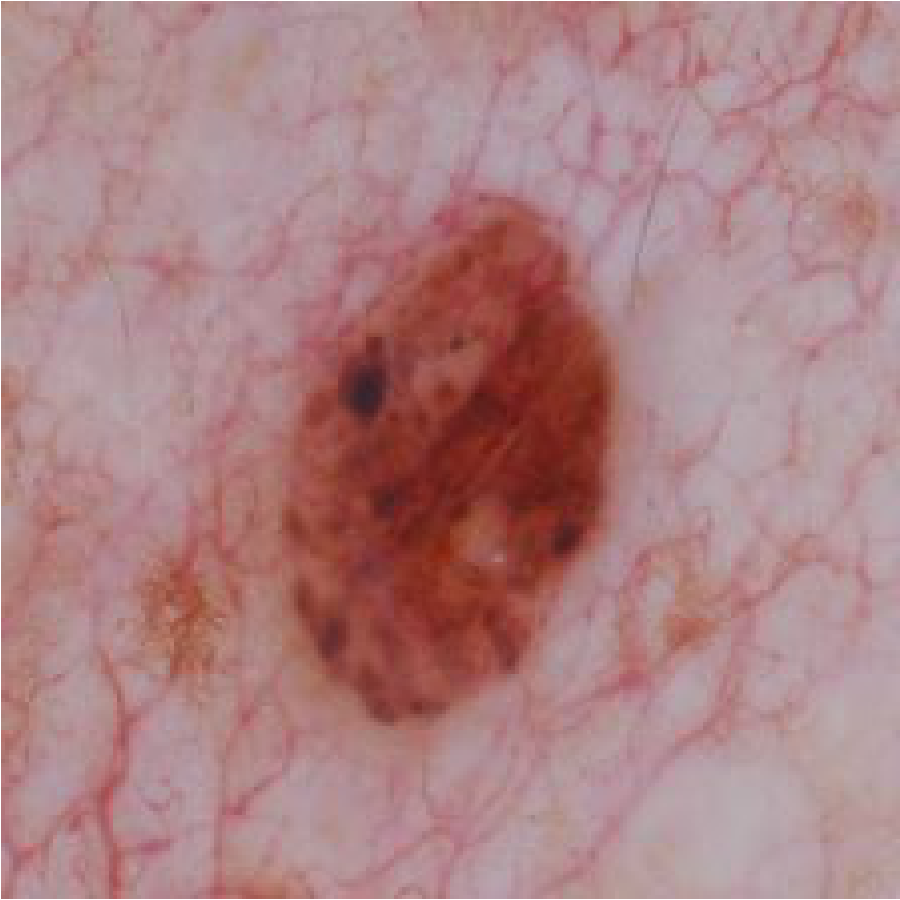}
		\caption{Img. $\Xvect_2$}
        \label{fig4:b1} 
	\end{subfigure}
\begin{subfigure}[H]{0.19\linewidth}
		\centering
  \includegraphics[height=2.3cm,width=2.3cm]{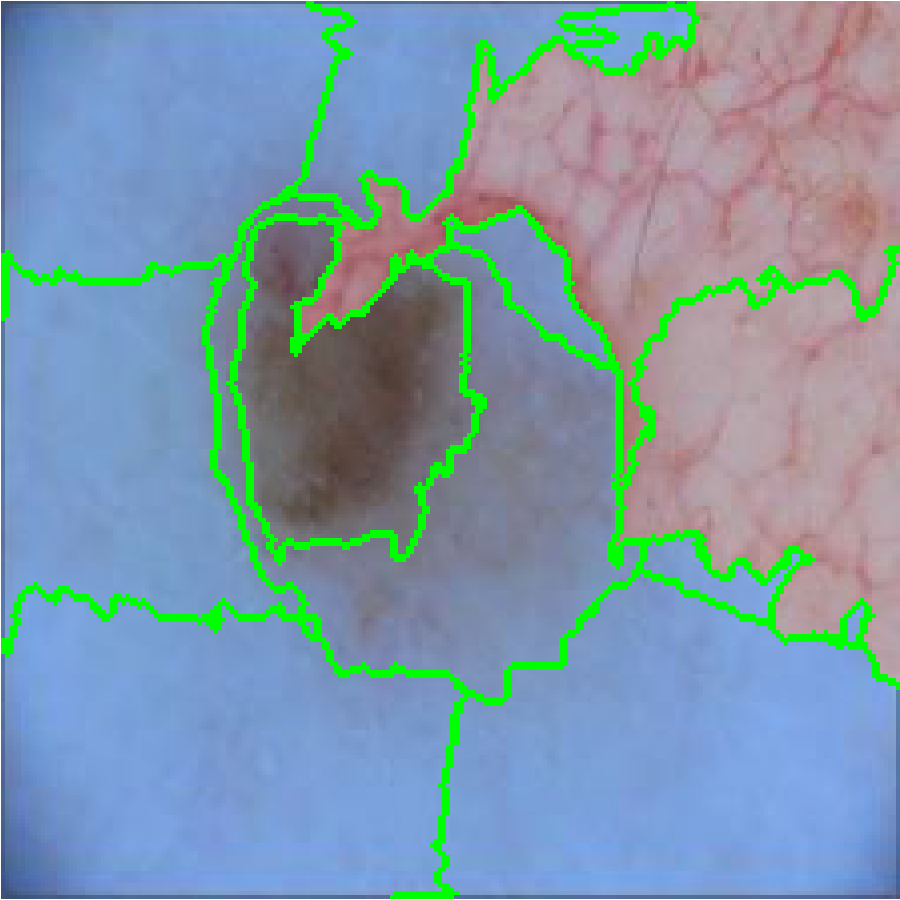}
		\caption{$l_{1}=l_{2}=10$}
        \label{fig4:c1} 
	\end{subfigure}
\begin{subfigure}[H]{0.19\linewidth}
		\centering
  \includegraphics[height=2.3cm,width=2.3cm]{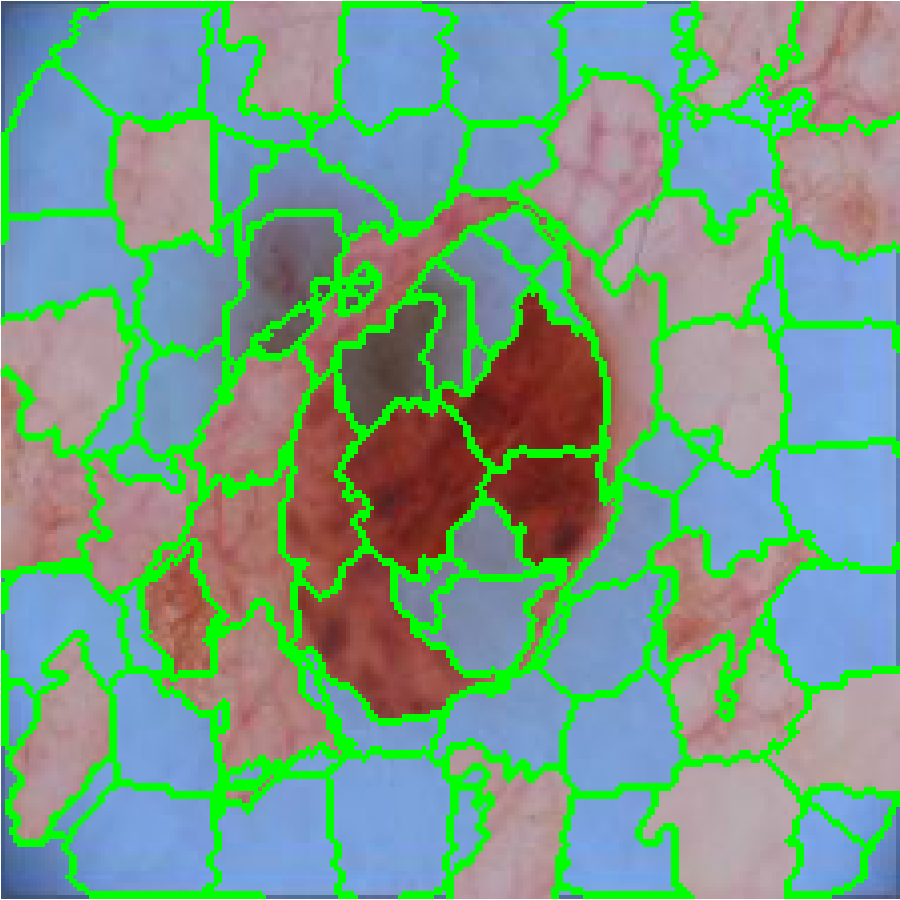}
		\caption{$l_{1}=l_{2}=60$}
        \label{fig4:d1} 
	\end{subfigure}
\begin{subfigure}[H]{0.19\linewidth}
		\centering
\includegraphics[height=2.3cm,width=2.3cm]{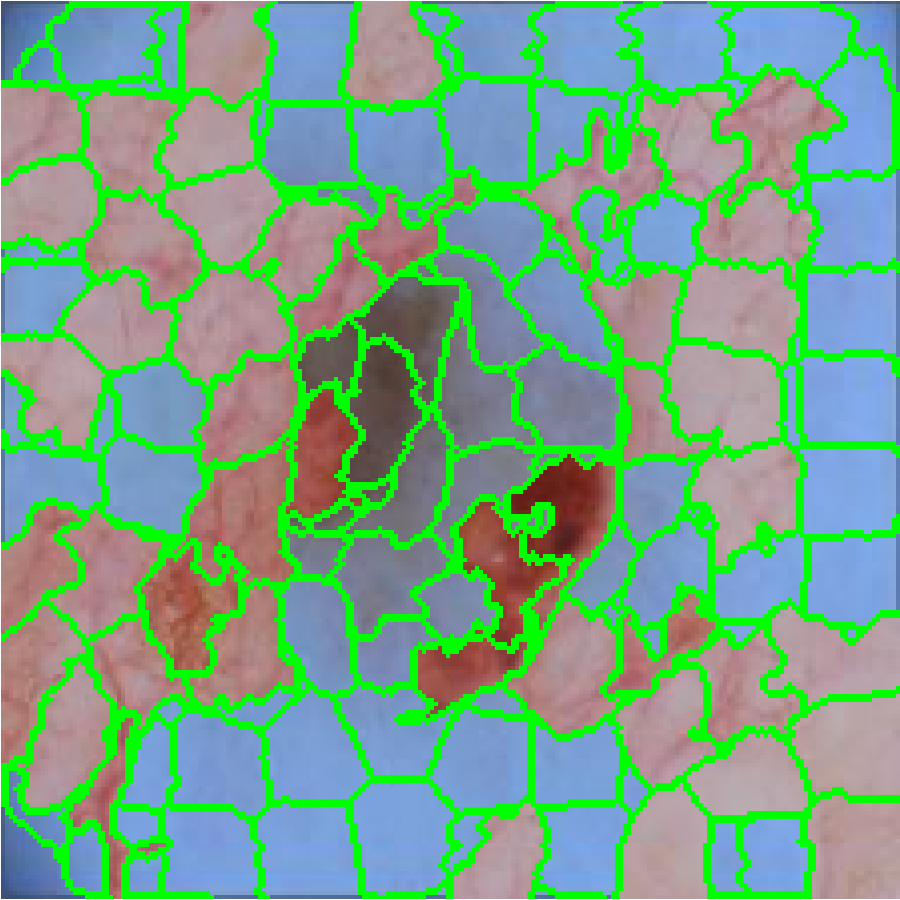}
		\caption{$l_{1}=l_{2}=100$}
        \label{fig4:e1} 
	\end{subfigure} 
 \caption{{(a)(b)(f)(g) Training images; (c)(d)(e)(h)(i)(j) Generated augmented images with outlined superpixel grids in green color.  The numbers of superpixels, $l_{1}$ and $l_{2}$ corresponding to either training image, are fixed here for visualization. }}
    \label{fig4}	
\end{figure*}

\begin{table}[h!]
\centering
\caption{\label{Table7}
Impact of the superpixel number $l$ on the ISIC 2017 Task 1 and the GlaS datasets using the UNet model.}
\setlength{\tabcolsep}{10mm}{
\resizebox{0.7\textwidth}{!}{
\begin{tabular}{c|rr}
\toprule
\multirow{2}{*}{\textbf{$l$}}  & \multicolumn{2}{c}{\textbf{ISIC 2017 T1}} \\
\cline{2-3}
& \textbf{DSC($\%$)$\uparrow$}  & \textbf{JAC($\%$)$\uparrow$} \\
\midrule
{$Constant: l=65$}  & $81.73\pm{0.83}$ & $72.93\pm{0.67}$ \\ 
\hline
$l \sim U(10,60)$  & $82.47\pm{0.62}$ & $73.15\pm{0.79}$ \\
$l \sim U(30,80)$  & $\bold{83.50\pm{0.67}}$ & $\bold{74.13\pm{0.51}}$  \\
$l \sim U(60,110)$  & $83.41\pm{0.87}$ & $73.89\pm{0.92}$  \\
$l \sim U(90,140)$  & $83.18\pm{0.68}$ & $73.55\pm{0.73}$ \\
\toprule
\multirow{2}{*}{\textbf{$l$}}  & \multicolumn{2}{c}{\textbf{GlaS}} \\
\cline{2-3}
& \textbf{DSC($\%$)$\uparrow$}  & \textbf{JAC($\%$)$\uparrow$} \\
\midrule
{$Constant: l=300$}  & $89.27\pm{0.70}$ & $82.26\pm{0.97}$ \\ 
\hline
$l \sim U(100,300)$  & $89.94\pm{0.48}$ & $83.04\pm{0.63}$ \\
$l \sim U(200,400)$  & $\bold{90.83\pm{0.90}}$ & $\bold{83.97\pm{1.36}}$ \\
$l \sim U(300,500)$  & $90.18\pm{1.53}$ & $83.29\pm{1.38}$ \\
$l \sim U(500,700)$  & $89.83\pm{1.87}$ & $82.96\pm{1.47}$ \\
\bottomrule
\end{tabular}}}
\end{table}


\begin{figure*}[t!]
      \centering
  \begin{subfigure}[H]{0.48\linewidth}
		\centering
  \includegraphics[height=4cm,width=5.5cm]{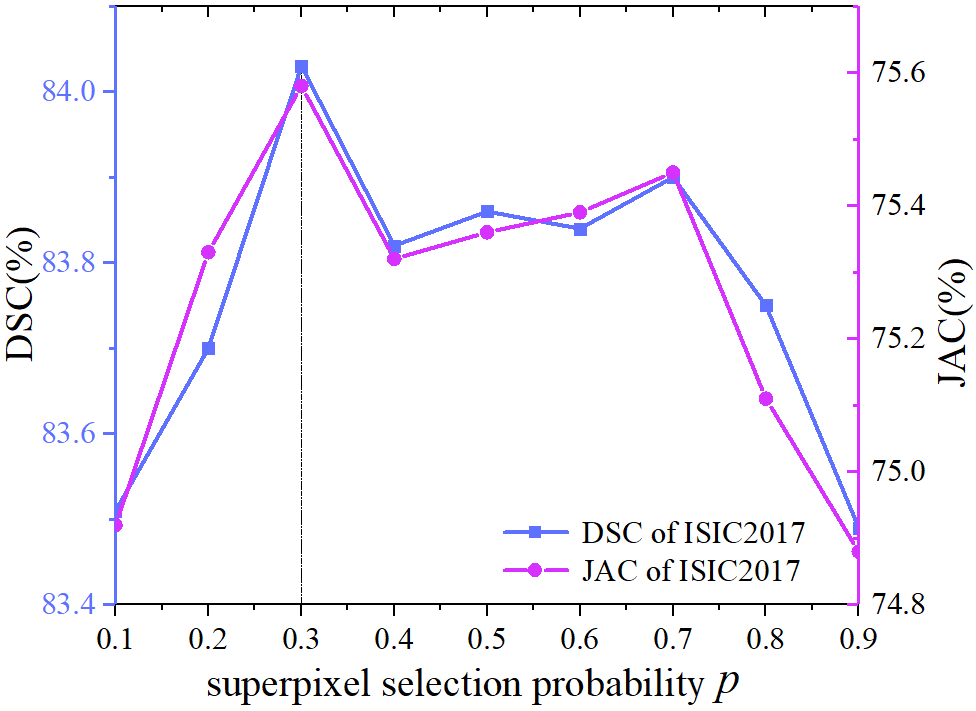}
        \label{fig5:a} 
	\end{subfigure} 
 \begin{subfigure}[H]{0.48\linewidth}
		\centering
  \includegraphics[height=4cm,width=5.5cm]{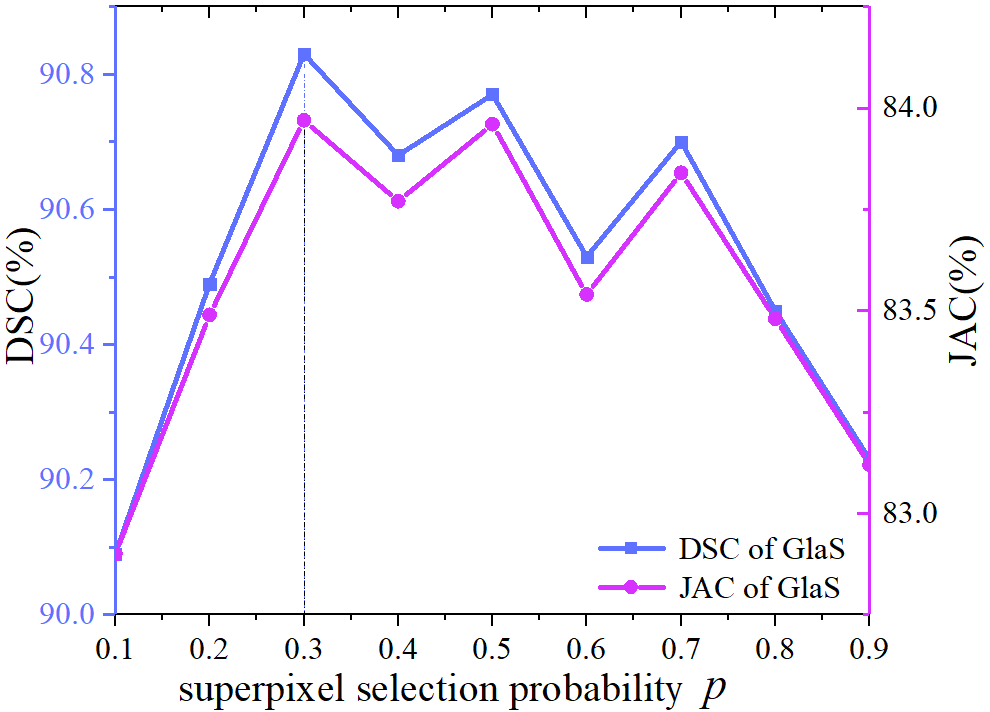}
        \label{fig5:b} 
	\end{subfigure}
 \caption{Performance changing with $p$, the selection probability of superpixels, on ISIC2017 Task 1 and Glas dataset using the UNet model. The best is achieved when $p=0.3$.}   
    \label{fig5}	
\end{figure*}

\textcolor{black}{In general, increasing the number of superpixel, $l$, leads to a finer granularity that provides less semantic information but more contour information.  Conversely, decreasing in the superpixel number results in a coarser granularity that provides less contour information but more semantic information.} \textcolor{black}{As can be seen in Figure \ref{fig4}, the superpixels in Figure \ref{fig4:c} and Figure \ref{fig4:c1} have a higher level of semantic information but lower capacity of contour information when the superpixel number, denoted by $l$, is relatively small}.  In Figures \ref{fig4:e} and \ref{fig4:e1}, the number of superpixels, $l$, is relatively large. Consequently, although the synthetic superpixel maps have detailed boundaries, they lack semantics. \textcolor{black}{For images with a large amount of fine-grained content or granularity, it is recommended to use a relatively large number of superpixels, where each superpixel is relatively small. Conversely, a small number of superpixels can also be used, with each superpixel being relatively large. In our method, the number of superpixels $l$ for each image is randomly determined according to a uniform distribution $l\sim U(l_{min}, l_{max})$ instead of setting it as constant. This is easier to tune and can increase the diversity of the augmentation space. It can also improve the robustness of our method.}  The quantitative analysis is shown in Table \ref{Table7}.

 \textcolor{black}{Higher values for compactness indicate that greater importance is attached to spatial proximity, leading to the formation of superpixel shapes that tend to be square or cubic. Lower values for compactness, on the other hand, indicate that greater importance is placed on color proximity, resulting in the formation of superpixel shapes that are irregular and have less regular edges. For medical CT and MRI images with low contrast or poor image quality, the compactness value is relatively low. Medical images with high contrast usually require higher compactness values, such as camera images of skin lesions.}

Our method involves the random selection of superpixels according to a Bernoulli distribution parameterized by $p$. The optimal hyperparameter value for $p$ is determined by the classical grid search technique. As can be seen in Figure \ref{fig5}, we obtain the best result when the probability of superpixel selection is set to $0.3$. In our evaluation, we set the probability of selecting a superpixel to $p=0.3$ for all datasets.

\begin{table}[h]
\centering
\caption{\textcolor{black}{\label{table9} Values of the main hyperparameters for each dataset.}}
\setlength{\tabcolsep}{10mm}{
\resizebox{0.95\textwidth}{!}{
\begin{tabular}{lc|ccc}
\toprule
  Datastes & Modality & $l \sim(l_{min}, l_{max})$ & $c$ & $p$ \\
\hline
GlaS & gland image under microscope & $l \sim U(200,400)$  & 10 & 0.3 \\ 
MoNuSeg & nuclear image under microscope &$l \sim U(300,500)$ & 10 & 0.3\\ 
ISIC 2017 T1 &camera image of skin lesion &$l \sim U(30,80)$  & 10 & 0.3\\ 
BraTS2018 &MRI image &$l \sim U(50,150)$  & 0.003 & 0.3\\ 
Synapse &CT image &$l \sim U(200,400)$  & 0.1 & 0.3\\ 
\bottomrule
\end{tabular}}}
\end{table}

\textcolor{black}{The hyperparameter values used for the generation and selection of superpixels are summarized in Table \ref{table9}. These values can be used as a reference and applied directly to medical image data of the same modality and type, so further tunning may be unnecessary.}

\subsection{Influence of Image Size on Performance}
\label{input_size}

\textcolor{black}{As shown in Table \ref{Table8}, larger image sizes can improve the performance of both the baseline results and the results obtained with the proposed HSMix. Since larger image sizes tend to preserve more details, this can help the models to learn finer features. This is particularly important in the context of medical images. In addition, larger image sizes can facilitate the creation of more accurate superpixel and saliency maps, which can improve the performance of our HSMix. Larger images make it easier, for example, to align superpixels with object boundaries. This is due to the higher resolution, which allows for more precise clustering around contours. The saliency maps derived from larger images show higher sensitivity to slight variations in color and intensity, resulting in higher accuracy. As can be seen in Table \ref{Table8}, our HSMix is able to achieve better performance when the input size is either $224\times{224}$ or $512\times{512}$. It is also important to note that larger image sizes can lead to increased memory usage, larger model sizes for transformer-based deep models, and longer training times.}

\begin{table*}[h!]
\centering
\caption{\label{Table8} Influence of Image Input Size on Performance}
\setlength{\tabcolsep}{1.5mm}{
\resizebox{0.9\textwidth}{!}{
\begin{tabular}{llcccc}
\toprule
\multirow{2}{*}{\textbf{Dataset}} & \multirow{2}{*}{\textbf{Method}}  & \multicolumn{2}{c}{\textbf{Image Size of 224}} & \multicolumn{2}{c}{\textbf{Image Size of 512}}\\ 
\cline{3-6} 

& & \textbf{DSC($\%$)$\uparrow$}  & \textbf{JAC($\%$)$\uparrow$} & \textbf{DSC($\%$)$\uparrow$}  & \textbf{JAC($\%$)$\uparrow$} \\
    \midrule
    \multirow{2}{*}{ISIC2017T1} & UNeXt \cite{valanarasu2022unext} &  $81.14\pm{0.98}$ & $72.71\pm{1.11}$ & $82.44\pm{0.87}$ & $73.62\pm{0.96}$\\
    & UNeXt + HSMix &  $\bold{82.74\pm{0.37}}$ & $\bold{73.91\pm{0.96}}$ & $\bold{83.56\pm{0.75}}$ & $\bold{74.21\pm{0.68}}$\\
    \midrule
    \multirow{2}{*}{MoNuSeg} & UNeXt \cite{valanarasu2022unext} &  $71.61\pm{1.31}$ & $56.24\pm{1.49}$ & $77.55\pm{0.38}$ & $63.73\pm{0.37}$\\
    & UNeXt + HSMix &  $\bold{72.29\pm{2.11}}$ & $\bold{56.84\pm{2.43}}$ & $\bold{77.93\pm{1.27}}$ & $\bold{64.28\pm{1.53}}$\\
    \midrule
    \multirow{2}{*}{GlaS} & UNeXt \cite{valanarasu2022unext} &  $82.00\pm{1.86}$ & $70.96\pm{2.44}$ & $85.83\pm{0.42}$ & $76.22\pm{0.59}$\\
    & UNeXt + HSMix &  $\bold{86.20\pm{0.88}}$ & $\bold{76.85\pm{1.23}}$ & $\bold{88.51\pm{0.47}}$ & $\bold{80.37\pm{0.65}}$\\

\bottomrule
\vspace{-1em}
\end{tabular}}}
\end{table*}

\subsection{Visualization and qualitative evaluation}
\label{comparisions}

\begin{figure}[t]
	\centering
	\begin{subfigure}[]{0.3\textwidth} {
    \centering
    \includegraphics[height=5.9cm,width=3.4cm]{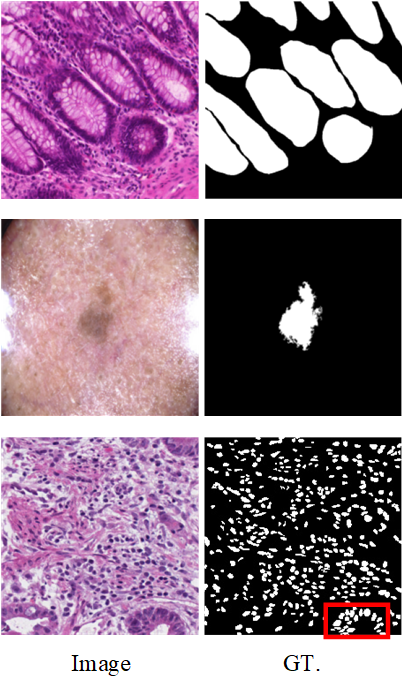}
    \caption{Images and GT masks.}
    \label{fig6:a}
    } 
    \end{subfigure}
	\begin{subfigure}[]{0.3\textwidth} {
    \centering    \includegraphics[height=5.9cm,width=3.4cm]{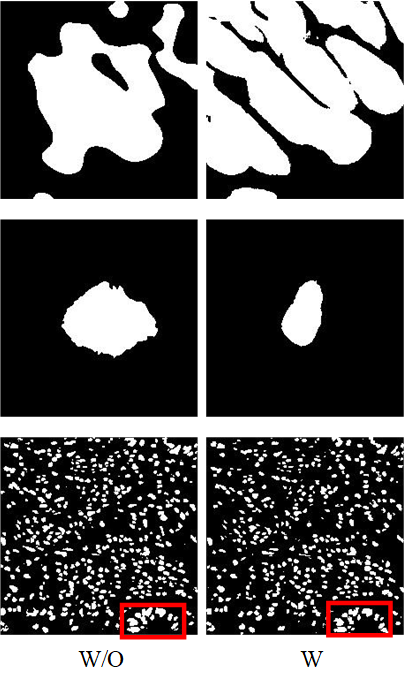}
    \caption{UNet}
    \label{fig6:b}
    }
    \end{subfigure}
	\begin{subfigure}[]{0.3\textwidth} {
    \centering    \includegraphics[height=5.9cm,width=3.4cm]{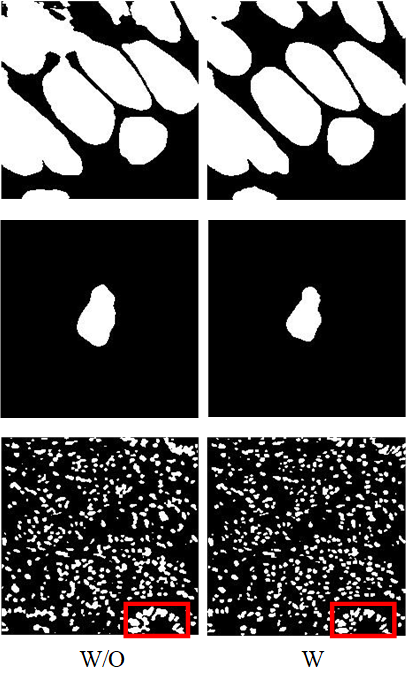}
    \caption{TransUnet}
    \label{fig6:c}
    }
    \end{subfigure}
    \caption{{Comparison of the predicted masks for binary medical segmentation tasks. (a) Images with ground truth segmentation masks. (b) Comparison of predictions with (W) or without (W/O) HSMix using UNet. (c) Comparison of predictions with (W) or without (W/O) HSMix using TransUnet. }}
    \label{fig6}
\end{figure}

\begin{figure}[h!]
	\centering
\includegraphics[scale=0.49]{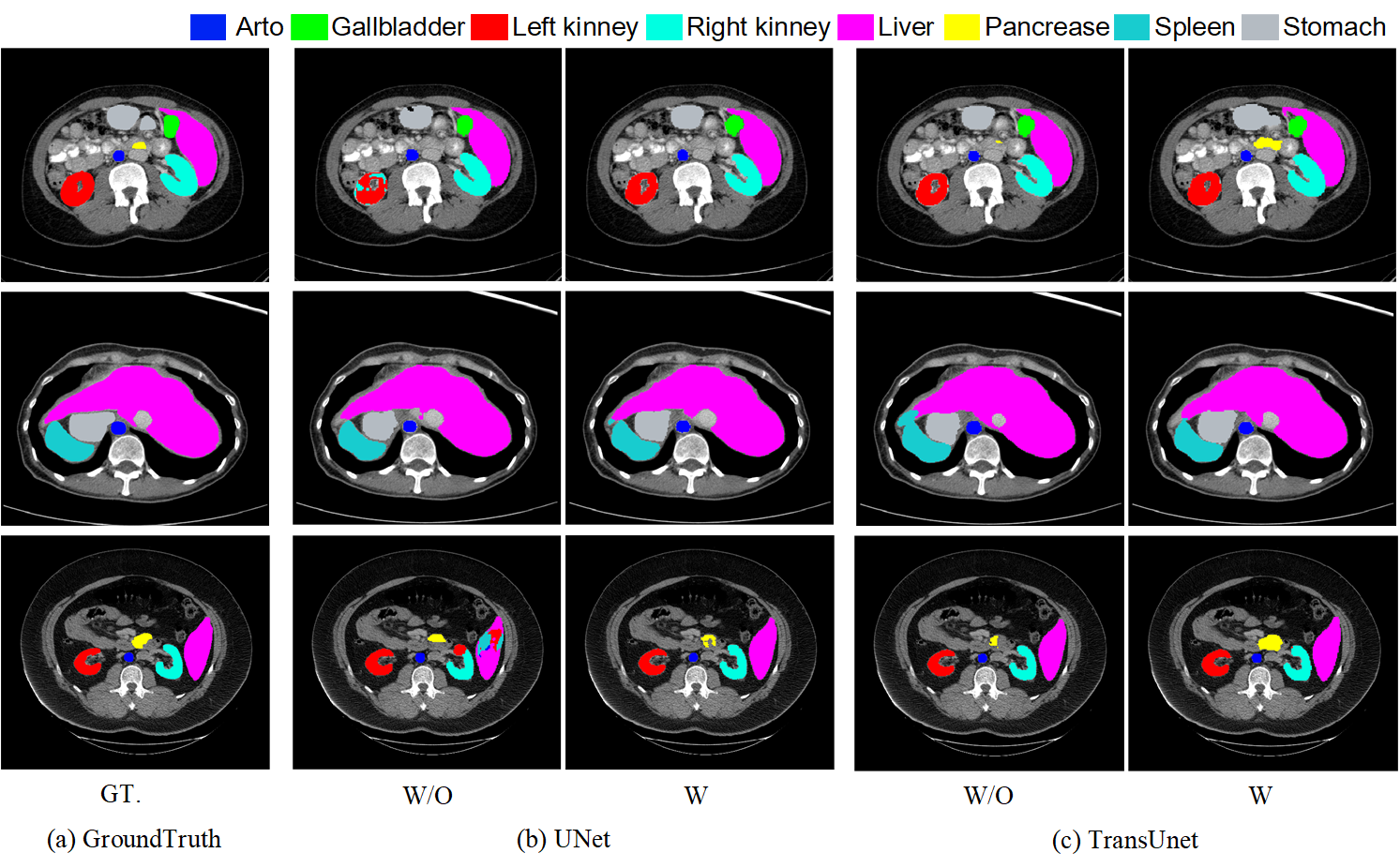}
    \caption{Comparison of the predicted masks for multi-label medical segmentation tasks of Synapse dataset. (a) Ground truth masks for multi-label CT images. (b) Comparison of predictions with (W) or without (W/O) HSMix using the UNet. (c) Comparison of predictions with (W) or without (W/O) HSMix using the TransUnet.}
    \label{fig7}
\end{figure}

Figures \ref{fig6} and  \ref{fig7} show a visual comparison of some predicted segmentation masks. Figure \ref{fig6} depicts the predicted masks of binary medical segmentation tasks. The image in the first row is from the GlaS test set. The results of the first row in Figure \ref{fig6} demonstrate that our proposed HSMix data augmentation method can improve the model's capacity to segment boundary regions more accurately during the inference phase. In the second row, for the segmentation of skin lesion in the ISIC 2017 T1, training with (W) the use of HSMix has fewer false-positive predictions compared to training without (W/O) the use of \textcolor{black}{the} HSMix method. The third row corresponds to the MoNuSeg dataset. As can be observed, training the models without (W/O) HSMix results in the prediction of individual small tissues as being connected.  Figure \ref{fig7} depicts the predicted masks of multi-label medical segmentation tasks for multi-organ CT images in the Synapse dataset. The first and third rows indicate that training with (W) HSMix has an advantage in predicting more true positives and filtering noise. The second row illustrates that training with HSMix can capture boundaries. In summary, training with (W) HSMix is more effective at filtering out noise and predicting contour boundaries than training without (W/O) this method.

\section{Conclusion}
\label{conclusion}

In this paper, we present HSMix, a novel approach \textcolor{black}{to} local image editing data augmentation that incorporates both hard and soft mixing for semantic segmentation of medical images. Our method fully utilizes superpixel regions and prior saliency information to ensure that important contour details are preserved in the augmented images while expanding the augmentation space with greater diversity. By \textcolor{black}{using} these elements, HSMix not only preserves the integrity of important features, but also improves the robustness of the augmented dataset. 
We also address the potential limitations arising from data scarcity, a common problem in medical imaging, and present a data augmentation approach to mitigate these challenges. The method we present is a plug-and-play solution \textcolor{black}{that is} model-independent and applicable to a range of medical imaging modalities, making it versatile and easy to integrate into existing workflows.
Extensive experimental evidence demonstrates the effectiveness of HSMix in various medical segmentation tasks. The results show significant improvements in segmentation accuracy and robustness, highlighting the potential of HSMix to improve medical image analysis.
\textcolor{black}{In the future, we plan to extend our research to the generation of synthetic medical data. This future focus will enable progress in both medical research and clinical practice. In medical research, we aim to support medical image analysis, provide a more comprehensive solution to the challenges of data scarcity and improve the overall capabilities of medical image segmentation. For clinical application, our method has the potential to improve the effectiveness of computer-aided diagnosis with segmentation for subsequent treatment, while mitigating the challenges associated with manual observation and segmentation.}


\begin{thebibliography}{10}

\bibitem{xu2021emfusion}
Han Xu and Jiayi Ma.
\newblock Emfusion: An unsupervised enhanced medical image fusion network.
\newblock {\em Information Fusion}, 76:177--186, 2021.

\bibitem{jha2020doubleu}
Debesh Jha, Michael~A Riegler, Dag Johansen, P{\aa}l Halvorsen, and
  H{\aa}vard~D Johansen.
\newblock Doubleu-net: A deep convolutional neural network for medical image
  segmentation.
\newblock In {\em 2020 IEEE 33rd International symposium on computer-based
  medical systems (CBMS)}, pages 558--564. IEEE, 2020.

\bibitem{qiu2022dwarfism}
Shi Qiu, Yi~Jin, Songhe Feng, Tao Zhou, and Yidong Li.
\newblock Dwarfism computer-aided diagnosis algorithm based on multimodal
  pyradiomics.
\newblock {\em Information Fusion}, 80:137--145, 2022.

\bibitem{ma2024segment}
Jun Ma, Yuting He, Feifei Li, Lin Han, Chenyu You, and Bo~Wang.
\newblock Segment anything in medical images.
\newblock {\em Nature Communications}, 15(1):654, 2024.

\bibitem{khan2021deep}
Muhammad~Zubair Khan, Mohan~Kumar Gajendran, Yugyung Lee, and Muazzam~A Khan.
\newblock Deep neural architectures for medical image semantic segmentation.
\newblock {\em IEEE Access}, 9:83002--83024, 2021.

\bibitem{jha2019resunet++}
Debesh Jha, Pia~H Smedsrud, Michael~A Riegler, Dag Johansen, Thomas De~Lange,
  P{\aa}l Halvorsen, and H{\aa}vard~D Johansen.
\newblock Resunet++: An advanced architecture for medical image segmentation.
\newblock In {\em 2019 IEEE international symposium on multimedia (ISM)}, pages
  225--2255. IEEE, 2019.

\bibitem{zhang2023dive}
Chuyan Zhang, Hao Zheng, and Yun Gu.
\newblock Dive into the details of self-supervised learning for medical image
  analysis.
\newblock {\em Medical Image Analysis}, 89:102879, 2023.

\bibitem{jiao2023learning}
Rushi Jiao, Yichi Zhang, Le~Ding, Bingsen Xue, Jicong Zhang, Rong Cai, and
  Cheng Jin.
\newblock Learning with limited annotations: a survey on deep semi-supervised
  learning for medical image segmentation.
\newblock {\em Computers in Biology and Medicine}, page 107840, 2023.

\bibitem{wang2021regularizing}
Yulin Wang, Gao Huang, Shiji Song, Xuran Pan, Yitong Xia, and Cheng Wu.
\newblock Regularizing deep networks with semantic data augmentation.
\newblock {\em IEEE Transactions on Pattern Analysis and Machine Intelligence},
  44(7):3733--3748, 2021.

\bibitem{goceri2023medical}
Evgin Goceri.
\newblock Medical image data augmentation: techniques, comparisons and
  interpretations.
\newblock {\em Artificial Intelligence Review}, pages 1--45, 2023.

\bibitem{goodfellow2020generative}
Ian Goodfellow, Jean Pouget-Abadie, Mehdi Mirza, Bing Xu, David Warde-Farley,
  Sherjil Ozair, Aaron Courville, and Yoshua Bengio.
\newblock Generative adversarial networks.
\newblock {\em Communications of the ACM}, 63(11):139--144, 2020.

\bibitem{zhang2018mixup}
Hongyi Zhang, Moustapha Cisse, Yann~N Dauphin, and David Lopez-Paz.
\newblock mixup: Beyond empirical risk minimization.
\newblock In {\em International Conference on Learning Representations}, 2018.

\bibitem{devries2017improved}
Terrance DeVries and Graham~W Taylor.
\newblock Improved regularization of convolutional neural networks with cutout.
\newblock {\em arXiv preprint arXiv:1708.04552}, 2017.

\bibitem{yun2019cutmix}
Sangdoo Yun, Dongyoon Han, Seong~Joon Oh, Sanghyuk Chun, Junsuk Choe, and
  Youngjoon Yoo.
\newblock Cutmix: Regularization strategy to train strong classifiers with
  localizable features.
\newblock In {\em Proceedings of the IEEE/CVF international conference on
  computer vision}, pages 6023--6032, 2019.

\bibitem{hammoudi2022superpixelgridmasks}
Karim Hammoudi, Adnane Cabani, Bouthaina Slika, Halim Benhabiles, Fadi
  Dornaika, and Mahmoud Melkemi.
\newblock Superpixelgridmasks data augmentation: Application to precision
  health and other real-world data.
\newblock {\em Journal of Healthcare Informatics Research}, 6(4):442--460,
  2022.

\bibitem{li2015automatic}
Wen Li, Fucang Jia, and Qingmao Hu.
\newblock Automatic segmentation of liver tumor in ct images with deep
  convolutional neural networks.
\newblock {\em Journal of Computer and Communications}, 3(11):146--151, 2015.

\bibitem{vivanti2015automatic}
R~Vivanti, A~Ephrat, L~Joskowicz, O~Karaaslan, N~Lev-Cohain, and J~Sosna.
\newblock Automatic liver tumor segmentation in follow-up ct studies using
  convolutional neural networks.
\newblock In {\em Proc. patch-based methods in medical image processing
  workshop}, volume~2, page~2, 2015.

\bibitem{menze2014multimodal}
Bjoern~H Menze, Andras Jakab, Stefan Bauer, Jayashree Kalpathy-Cramer, Keyvan
  Farahani, Justin Kirby, Yuliya Burren, Nicole Porz, Johannes Slotboom, Roland
  Wiest, et~al.
\newblock The multimodal brain tumor image segmentation benchmark (brats).
\newblock {\em IEEE transactions on medical imaging}, 34(10):1993--2024, 2014.

\bibitem{cherukuri2017learning}
Venkateswararao Cherukuri, Peter Ssenyonga, Benjamin~C Warf, Abhaya~V Kulkarni,
  Vishal Monga, and Steven~J Schiff.
\newblock Learning based segmentation of ct brain images: application to
  postoperative hydrocephalic scans.
\newblock {\em IEEE transactions on biomedical engineering}, 65(8):1871--1884,
  2017.

\bibitem{cheng2013superpixel}
Jun Cheng, Jiang Liu, Yanwu Xu, Fengshou Yin, Damon Wing~Kee Wong, Ngan-Meng
  Tan, Dacheng Tao, Ching-Yu Cheng, Tin Aung, and Tien~Yin Wong.
\newblock Superpixel classification based optic disc and optic cup segmentation
  for glaucoma screening.
\newblock {\em IEEE transactions on medical imaging}, 32(6):1019--1032, 2013.

\bibitem{fu2018joint}
Huazhu Fu, Jun Cheng, Yanwu Xu, Damon Wing~Kee Wong, Jiang Liu, and Xiaochun
  Cao.
\newblock Joint optic disc and cup segmentation based on multi-label deep
  network and polar transformation.
\newblock {\em IEEE transactions on medical imaging}, 37(7):1597--1605, 2018.

\bibitem{song2017dual}
Tzu-Hsi Song, Victor Sanchez, Hesham EIDaly, and Nasir~M Rajpoot.
\newblock Dual-channel active contour model for megakaryocytic cell
  segmentation in bone marrow trephine histology images.
\newblock {\em IEEE transactions on biomedical engineering}, 64(12):2913--2923,
  2017.

\bibitem{wei2020mitoem}
Donglai Wei, Zudi Lin, Daniel Franco-Barranco, Nils Wendt, Xingyu Liu, Wenjie
  Yin, Xin Huang, Aarush Gupta, Won-Dong Jang, Xueying Wang, et~al.
\newblock Mitoem dataset: Large-scale 3d mitochondria instance segmentation
  from em images.
\newblock In {\em International Conference on Medical Image Computing and
  Computer-Assisted Intervention}, pages 66--76. Springer, 2020.

\bibitem{wang2017central}
Shuo Wang, Mu~Zhou, Zaiyi Liu, Zhenyu Liu, Dongsheng Gu, Yali Zang, Di~Dong,
  Olivier Gevaert, and Jie Tian.
\newblock Central focused convolutional neural networks: Developing a
  data-driven model for lung nodule segmentation.
\newblock {\em Medical image analysis}, 40:172--183, 2017.

\bibitem{wu2020cf}
Fuping Wu and Xiahai Zhuang.
\newblock Cf distance: a new domain discrepancy metric and application to
  explicit domain adaptation for cross-modality cardiac image segmentation.
\newblock {\em IEEE Transactions on Medical Imaging}, 39(12):4274--4285, 2020.

\bibitem{yuan2017automatic}
Yading Yuan, Ming Chao, and Yeh-Chi Lo.
\newblock Automatic skin lesion segmentation using deep fully convolutional
  networks with jaccard distance.
\newblock {\em IEEE transactions on medical imaging}, 36(9):1876--1886, 2017.

\bibitem{litjens2017survey}
Geert Litjens, Thijs Kooi, Babak~Ehteshami Bejnordi, Arnaud Arindra~Adiyoso
  Setio, Francesco Ciompi, Mohsen Ghafoorian, Jeroen~Awm Van Der~Laak, Bram
  Van~Ginneken, and Clara~I S{\'a}nchez.
\newblock A survey on deep learning in medical image analysis.
\newblock {\em Medical image analysis}, 42:60--88, 2017.

\bibitem{minaee2021image}
Shervin Minaee, Yuri Boykov, Fatih Porikli, Antonio Plaza, Nasser Kehtarnavaz,
  and Demetri Terzopoulos.
\newblock Image segmentation using deep learning: A survey.
\newblock {\em IEEE transactions on pattern analysis and machine intelligence},
  44(7):3523--3542, 2021.

\bibitem{antonelli2022medical}
Michela Antonelli, Annika Reinke, Spyridon Bakas, Keyvan Farahani, Annette
  Kopp-Schneider, Bennett~A Landman, Geert Litjens, Bjoern Menze, Olaf
  Ronneberger, Ronald~M Summers, et~al.
\newblock The medical segmentation decathlon.
\newblock {\em Nature communications}, 13(1):4128, 2022.

\bibitem{ronneberger2015u}
Olaf Ronneberger, Philipp Fischer, and Thomas Brox.
\newblock U-net: Convolutional networks for biomedical image segmentation.
\newblock In {\em Medical image computing and computer-assisted
  intervention--MICCAI 2015: 18th international conference, Munich, Germany,
  October 5-9, 2015, proceedings, part III 18}, pages 234--241. Springer, 2015.

\bibitem{oktay2018attention}
Ozan Oktay, Jo~Schlemper, Loic~Le Folgoc, Matthew Lee, Mattias Heinrich,
  Kazunari Misawa, Kensaku Mori, Steven McDonagh, Nils~Y Hammerla, Bernhard
  Kainz, et~al.
\newblock Attention u-net: Learning where to look for the pancreas.
\newblock {\em arXiv preprint arXiv:1804.03999}, 2018.

\bibitem{milletari2016v}
Fausto Milletari, Nassir Navab, and Seyed-Ahmad Ahmadi.
\newblock V-net: Fully convolutional neural networks for volumetric medical
  image segmentation.
\newblock In {\em 2016 fourth international conference on 3D vision (3DV)},
  pages 565--571. Ieee, 2016.

\bibitem{zhang2018road}
Zhengxin Zhang, Qingjie Liu, and Yunhong Wang.
\newblock Road extraction by deep residual u-net.
\newblock {\em IEEE Geoscience and Remote Sensing Letters}, 15(5):749--753,
  2018.

\bibitem{lin2022survey}
Tianyang Lin, Yuxin Wang, Xiangyang Liu, and Xipeng Qiu.
\newblock A survey of transformers.
\newblock {\em AI open}, 3:111--132, 2022.

\bibitem{bazi2021vision}
Yakoub Bazi, Laila Bashmal, Mohamad M~Al Rahhal, Reham~Al Dayil, and Naif~Al
  Ajlan.
\newblock Vision transformers for remote sensing image classification.
\newblock {\em Remote Sensing}, 13(3):516, 2021.

\bibitem{xie2023maester}
Ronald Xie, Kuan Pang, Gary~D Bader, and Bo~Wang.
\newblock Maester: masked autoencoder guided segmentation at pixel resolution
  for accurate, self-supervised subcellular structure recognition.
\newblock In {\em Proceedings of the IEEE/CVF Conference on Computer Vision and
  Pattern Recognition}, pages 3292--3301, 2023.

\bibitem{tang2022self}
Yucheng Tang, Dong Yang, Wenqi Li, Holger~R Roth, Bennett Landman, Daguang Xu,
  Vishwesh Nath, and Ali Hatamizadeh.
\newblock Self-supervised pre-training of swin transformers for 3d medical
  image analysis.
\newblock In {\em Proceedings of the IEEE/CVF conference on computer vision and
  pattern recognition}, pages 20730--20740, 2022.

\bibitem{xie2020pgl}
Yutong Xie, Jianpeng Zhang, Zehui Liao, Yong Xia, and Chunhua Shen.
\newblock Pgl: Prior-guided local self-supervised learning for 3d medical image
  segmentation.
\newblock {\em arXiv preprint arXiv:2011.12640}, 2020.

\bibitem{chaitanya2023local}
Krishna Chaitanya, Ertunc Erdil, Neerav Karani, and Ender Konukoglu.
\newblock Local contrastive loss with pseudo-label based self-training for
  semi-supervised medical image segmentation.
\newblock {\em Medical image analysis}, 87:102792, 2023.

\bibitem{shurrab2022self}
Saeed Shurrab and Rehab Duwairi.
\newblock Self-supervised learning methods and applications in medical imaging
  analysis: A survey.
\newblock {\em PeerJ Computer Science}, 8:e1045, 2022.

\bibitem{ouyang2022self}
Cheng Ouyang, Carlo Biffi, Chen Chen, Turkay Kart, Huaqi Qiu, and Daniel
  Rueckert.
\newblock Self-supervised learning for few-shot medical image segmentation.
\newblock {\em IEEE Transactions on Medical Imaging}, 41(7):1837--1848, 2022.

\bibitem{luo2021semi}
Xiangde Luo, Jieneng Chen, Tao Song, and Guotai Wang.
\newblock Semi-supervised medical image segmentation through dual-task
  consistency.
\newblock In {\em Proceedings of the AAAI conference on artificial
  intelligence}, volume~35, pages 8801--8809, 2021.

\bibitem{bortsova2019semi}
Gerda Bortsova, Florian Dubost, Laurens Hogeweg, Ioannis Katramados, and
  Marleen De~Bruijne.
\newblock Semi-supervised medical image segmentation via learning consistency
  under transformations.
\newblock In {\em Medical Image Computing and Computer Assisted
  Intervention--MICCAI 2019: 22nd International Conference, Shenzhen, China,
  October 13--17, 2019, Proceedings, Part VI 22}, pages 810--818. Springer,
  2019.

\bibitem{chen2019multi}
Shuai Chen, Gerda Bortsova, Antonio Garc{\'\i}a-Uceda~Ju{\'a}rez, Gijs
  Van~Tulder, and Marleen De~Bruijne.
\newblock Multi-task attention-based semi-supervised learning for medical image
  segmentation.
\newblock In {\em Medical Image Computing and Computer Assisted
  Intervention--MICCAI 2019: 22nd International Conference, Shenzhen, China,
  October 13--17, 2019, Proceedings, Part III 22}, pages 457--465. Springer,
  2019.

\bibitem{bai2017semi}
Wenjia Bai, Ozan Oktay, Matthew Sinclair, Hideaki Suzuki, Martin Rajchl,
  Giacomo Tarroni, Ben Glocker, Andrew King, Paul~M Matthews, and Daniel
  Rueckert.
\newblock Semi-supervised learning for network-based cardiac mr image
  segmentation.
\newblock In {\em Medical Image Computing and Computer-Assisted Intervention-
  MICCAI 2017: 20th International Conference, Quebec City, QC, Canada,
  September 11-13, 2017, Proceedings, Part II 20}, pages 253--260. Springer,
  2017.

\bibitem{sun2024semi}
Huajun Sun, Jia Wei, Wenguang Yuan, and Rui Li.
\newblock Semi-supervised multi-modal medical image segmentation with unified
  translation.
\newblock {\em Computers in Biology and Medicine}, page 108570, 2024.

\bibitem{xie2022simmim}
Zhenda Xie, Zheng Zhang, Yue Cao, Yutong Lin, Jianmin Bao, Zhuliang Yao,
  Qi~Dai, and Han Hu.
\newblock Simmim: A simple framework for masked image modeling.
\newblock In {\em Proceedings of the IEEE/CVF Conference on Computer Vision and
  Pattern Recognition}, pages 9653--9663, 2022.

\bibitem{bai2023bidirectional}
Yunhao Bai, Duowen Chen, Qingli Li, Wei Shen, and Yan Wang.
\newblock Bidirectional copy-paste for semi-supervised medical image
  segmentation.
\newblock In {\em Proceedings of the IEEE/CVF Conference on Computer Vision and
  Pattern Recognition}, pages 11514--11524, 2023.

\bibitem{haghighi2024self}
Fatemeh Haghighi, Mohammad Reza~Hosseinzadeh Taher, Michael~B Gotway, and
  Jianming Liang.
\newblock Self-supervised learning for medical image analysis: Discriminative,
  restorative, or adversarial?
\newblock {\em Medical Image Analysis}, page 103086, 2024.

\bibitem{weng2024semi}
Ying Weng, Yiming Zhang, Wenxin Wang, and Tom Dening.
\newblock Semi-supervised information fusion for medical image analysis: Recent
  progress and future perspectives.
\newblock {\em Information Fusion}, page 102263, 2024.

\bibitem{chaitanya2021semi}
Krishna Chaitanya, Neerav Karani, Christian~F Baumgartner, Ertunc Erdil, Anton
  Becker, Olivio Donati, and Ender Konukoglu.
\newblock Semi-supervised task-driven data augmentation for medical image
  segmentation.
\newblock {\em Medical Image Analysis}, 68:101934, 2021.

\bibitem{hussain2017differential}
Zeshan Hussain, Francisco Gimenez, Darvin Yi, and Daniel Rubin.
\newblock Differential data augmentation techniques for medical imaging
  classification tasks.
\newblock In {\em AMIA annual symposium proceedings}, volume 2017, page 979.
  American Medical Informatics Association, 2017.

\bibitem{sharma2017automatic}
Kanishka Sharma, Christian Rupprecht, Anna Caroli, Maria~Carolina Aparicio,
  Andrea Remuzzi, Maximilian Baust, and Nassir Navab.
\newblock Automatic segmentation of kidneys using deep learning for total
  kidney volume quantification in autosomal dominant polycystic kidney disease.
\newblock {\em Scientific reports}, 7(1):2049, 2017.

\bibitem{alnazer2021recent}
Israa Alnazer, Pascal Bourdon, Thierry Urruty, Omar Falou, Mohamad Khalil,
  Ahmad Shahin, and Christine Fernandez-Maloigne.
\newblock Recent advances in medical image processing for the evaluation of
  chronic kidney disease.
\newblock {\em Medical Image Analysis}, 69:101960, 2021.

\bibitem{khan2021brain}
Amjad~Rehman Khan, Siraj Khan, Majid Harouni, Rashid Abbasi, Sajid Iqbal, and
  Zahid Mehmood.
\newblock Brain tumor segmentation using k-means clustering and deep learning
  with synthetic data augmentation for classification.
\newblock {\em Microscopy Research and Technique}, 84(7):1389--1399, 2021.

\bibitem{abdelhalim2021data}
Ibrahim Saad~Aly Abdelhalim, Mamdouh~Farouk Mohamed, and Yousef~Bassyouni
  Mahdy.
\newblock Data augmentation for skin lesion using self-attention based
  progressive generative adversarial network.
\newblock {\em Expert Systems with Applications}, 165:113922, 2021.

\bibitem{ding2021high}
Saisai Ding, Jian Zheng, Zhaobang Liu, Yanyan Zheng, Yanmei Chen, Xiaomin Xu,
  Jia Lu, and Jing Xie.
\newblock High-resolution dermoscopy image synthesis with conditional
  generative adversarial networks.
\newblock {\em Biomedical Signal Processing and Control}, 64:102224, 2021.

\bibitem{frid2018gan}
Maayan Frid-Adar, Idit Diamant, Eyal Klang, Michal Amitai, Jacob Goldberger,
  and Hayit Greenspan.
\newblock Gan-based synthetic medical image augmentation for increased cnn
  performance in liver lesion classification.
\newblock {\em Neurocomputing}, 321:321--331, 2018.

\bibitem{toikkanen2021resgan}
Miika Toikkanen, Doyoung Kwon, and Minho Lee.
\newblock Resgan: Intracranial hemorrhage segmentation with residuals of
  synthetic brain ct scans.
\newblock In {\em Medical Image Computing and Computer Assisted
  Intervention--MICCAI 2021: 24th International Conference, Strasbourg, France,
  September 27--October 1, 2021, Proceedings, Part I 24}, pages 400--409.
  Springer, 2021.

\bibitem{han2019combining}
Changhee Han, Leonardo Rundo, Ryosuke Araki, Yudai Nagano, Yujiro Furukawa,
  Giancarlo Mauri, Hideki Nakayama, and Hideaki Hayashi.
\newblock Combining noise-to-image and image-to-image gans: Brain mr image
  augmentation for tumor detection.
\newblock {\em Ieee Access}, 7:156966--156977, 2019.

\bibitem{lei2020skin}
Baiying Lei, Zaimin Xia, Feng Jiang, Xudong Jiang, Zongyuan Ge, Yanwu Xu, Jing
  Qin, Siping Chen, Tianfu Wang, and Shuqiang Wang.
\newblock Skin lesion segmentation via generative adversarial networks with
  dual discriminators.
\newblock {\em Medical Image Analysis}, 64:101716, 2020.

\bibitem{zhao2018synthesizing}
He~Zhao, Huiqi Li, Sebastian Maurer-Stroh, and Li~Cheng.
\newblock Synthesizing retinal and neuronal images with generative adversarial
  nets.
\newblock {\em Medical image analysis}, 49:14--26, 2018.

\bibitem{wang2022data}
Yu~Wang, Yarong Ji, and Hongbing Xiao.
\newblock A data augmentation method for fully automatic brain tumor
  segmentation.
\newblock {\em Computers in Biology and Medicine}, 149:106039, 2022.

\bibitem{zhang2022self}
Jiapeng Zhang.
\newblock Self-pretrained v-net based on pcrl for abdominal organ segmentation.
\newblock In {\em MICCAI Challenge on Fast and Low-Resource Semi-supervised
  Abdominal Organ Segmentation}, pages 260--269. Springer, 2022.

\bibitem{basaran2023lesionmix}
Berke~Doga Basaran, Weitong Zhang, Mengyun Qiao, Bernhard Kainz, Paul~M
  Matthews, and Wenjia Bai.
\newblock Lesionmix: A lesion-level data augmentation method for medical image
  segmentation.
\newblock In {\em International Conference on Medical Image Computing and
  Computer-Assisted Intervention}, pages 73--83. Springer, 2023.

\bibitem{achanta2012slic}
Radhakrishna Achanta, Appu Shaji, Kevin Smith, Aurelien Lucchi, Pascal Fua, and
  Sabine S{\"u}sstrunk.
\newblock Slic superpixels compared to state-of-the-art superpixel methods.
\newblock {\em IEEE transactions on pattern analysis and machine intelligence},
  34(11):2274--2282, 2012.

\bibitem{wang2023autosmim}
Zhonghua Wang, Junyan Lyu, and Xiaoying Tang.
\newblock Autosmim: Automatic superpixel-based masked image modeling for skin
  lesion segmentation.
\newblock {\em IEEE Transactions on Medical Imaging}, 2023.

\bibitem{accion2020dual}
{\'A}lvaro Acci{\'o}n, Francisco Arg{\"u}ello, and Dora~B Heras.
\newblock Dual-window superpixel data augmentation for hyperspectral image
  classification.
\newblock {\em Applied Sciences}, 10(24):8833, 2020.

\bibitem{dornaika2023object}
Fadi Dornaika, D~Sun, Karim Hammoudi, Jinan Charafeddine, Adnane Cabani, and
  C~Zhang.
\newblock Object-centric contour-aware data augmentation using superpixels of
  varying granularity.
\newblock {\em Pattern Recognition}, 139:109481, 2023.

\bibitem{dornaika2023lgcoamix}
Fadi Dornaika and Danyang Sun.
\newblock Lgcoamix: Local and global context-and-object-part-aware
  superpixel-based data augmentation for deep visual recognition.
\newblock {\em IEEE Transactions on Image Processing}, 33:205--215, 2023.

\bibitem{franchi2021robust}
Gianni Franchi, Nacim Belkhir, Mai~Lan Ha, Yufei Hu, Andrei Bursuc, Volker
  Blanz, and Angela Yao.
\newblock Robust semantic segmentation with superpixel-mix.
\newblock In {\em Proceedings of the British Machine Vision Conference}, 2021.

\bibitem{zhang2019spda}
Yizhe Zhang, Lin Yang, Hao Zheng, Peixian Liang, Colleen Mangold, Raquel~G
  Loreto, David~P Hughes, and Danny~Z Chen.
\newblock Spda: Superpixel-based data augmentation for biomedical image
  segmentation.
\newblock In {\em International Conference on Medical Imaging with Deep
  Learning}, pages 572--587. PMLR, 2019.

\bibitem{tan2019efficientnet}
Mingxing Tan and Quoc Le.
\newblock Efficientnet: Rethinking model scaling for convolutional neural
  networks.
\newblock In {\em International conference on machine learning}, pages
  6105--6114. PMLR, 2019.

\bibitem{he2016deep}
Kaiming He, Xiangyu Zhang, Shaoqing Ren, and Jian Sun.
\newblock Deep residual learning for image recognition.
\newblock In {\em Proceedings of the IEEE conference on computer vision and
  pattern recognition}, pages 770--778, 2016.

\bibitem{chen2018encoder}
Liang-Chieh Chen, Yukun Zhu, George Papandreou, Florian Schroff, and Hartwig
  Adam.
\newblock Encoder-decoder with atrous separable convolution for semantic image
  segmentation.
\newblock In {\em Proceedings of the European conference on computer vision
  (ECCV)}, pages 801--818, 2018.

\bibitem{chen2021transunet}
Jieneng Chen, Yongyi Lu, Qihang Yu, Xiangde Luo, Ehsan Adeli, Yan Wang, Le~Lu,
  Alan~L Yuille, and Yuyin Zhou.
\newblock Transunet: Transformers make strong encoders for medical image
  segmentation.
\newblock {\em arXiv preprint arXiv:2102.04306}, 2021.

\bibitem{heidari2023hiformer}
Moein Heidari, Amirhossein Kazerouni, Milad Soltany, Reza Azad,
  Ehsan~Khodapanah Aghdam, Julien Cohen-Adad, and Dorit Merhof.
\newblock Hiformer: Hierarchical multi-scale representations using transformers
  for medical image segmentation.
\newblock In {\em Proceedings of the IEEE/CVF Winter Conference on Applications
  of Computer Vision}, pages 6202--6212, 2023.

\bibitem{valanarasu2022unext}
Jeya Maria~Jose Valanarasu and Vishal~M Patel.
\newblock Unext: Mlp-based rapid medical image segmentation network.
\newblock In {\em International Conference on Medical Image Computing and
  Computer-Assisted Intervention}, pages 23--33. Springer, 2022.

\bibitem{codella2018skin}
Noel~CF Codella, David Gutman, M~Emre Celebi, Brian Helba, Michael~A Marchetti,
  Stephen~W Dusza, Aadi Kalloo, Konstantinos Liopyris, Nabin Mishra, Harald
  Kittler, et~al.
\newblock Skin lesion analysis toward melanoma detection: A challenge at the
  2017 international symposium on biomedical imaging.
\newblock In {\em 2018 IEEE 15th international symposium on biomedical imaging
  (ISBI 2018)}, pages 168--172. IEEE, 2018.

\bibitem{sirinukunwattana2017gland}
Korsuk Sirinukunwattana, Josien~PW Pluim, Hao Chen, Xiaojuan Qi, Pheng-Ann
  Heng, Yun~Bo Guo, Li~Yang Wang, Bogdan~J Matuszewski, Elia Bruni, Urko
  Sanchez, et~al.
\newblock Gland segmentation in colon histology images: The glas challenge
  contest.
\newblock {\em Medical image analysis}, 35:489--502, 2017.

\bibitem{kumar2019multi}
Neeraj Kumar, Ruchika Verma, Deepak Anand, Yanning Zhou, Omer~Fahri Onder,
  Efstratios Tsougenis, Hao Chen, Pheng-Ann Heng, Jiahui Li, Zhiqiang Hu,
  et~al.
\newblock A multi-organ nucleus segmentation challenge.
\newblock {\em IEEE Transactions on medical imaging}, 39(5):1380--1391, 2019.

\bibitem{kermi2019deep}
Adel Kermi, Issam Mahmoudi, and Mohamed~Tarek Khadir.
\newblock Deep convolutional neural networks using u-net for automatic brain
  tumor segmentation in multimodal mri volumes.
\newblock In {\em Brainlesion: Glioma, Multiple Sclerosis, Stroke and Traumatic
  Brain Injuries: 4th International Workshop, BrainLes 2018, Held in
  Conjunction with MICCAI 2018, Granada, Spain, September 16, 2018, Revised
  Selected Papers, Part II 4}, pages 37--48. Springer, 2019.

\bibitem{bi2019improving}
Lei Bi, Dagan Feng, Michael Fulham, and Jinman Kim.
\newblock Improving skin lesion segmentation via stacked adversarial learning.
\newblock In {\em 2019 IEEE 16Th International symposium on biomedical imaging
  (ISBI 2019)}, pages 1100--1103. IEEE, 2019.

\bibitem{sun2024lcamix}
D~Sun, F~Dornaika, and J~Charafeddine.
\newblock Lcamix: Local-and-contour aware grid mixing based data augmentation
  for medical image segmentation.
\newblock {\em Information Fusion}, 110:102484, 2024.

\bibitem{sun2024data}
D~Sun and F~Dornaika.
\newblock Data augmentation for deep visual recognition using superpixel based
  pairwise image fusion.
\newblock {\em Information Fusion}, page 102308, 2024.

\bibitem{kang2023guidedmixup}
Minsoo Kang and Suhyun Kim.
\newblock Guidedmixup: an efficient mixup strategy guided by saliency maps.
\newblock In {\em Proceedings of the AAAI Conference on Artificial
  Intelligence}, volume~37, pages 1096--1104, 2023.

\end{thebibliography}

\end{document}